%% file: coling_latex.tex
\pdfoutput=1

\documentclass[11pt]{article}

\usepackage[final]{coling}

\usepackage{times}
\usepackage{latexsym}

\usepackage[T2A,T1]{fontenc}
\usepackage[utf8]{inputenc}
\usepackage[greek,russian,english]{babel}

\usepackage{microtype}
\usepackage{inconsolata}
\usepackage{graphicx}
\usepackage{microtype}
\usepackage{amsmath}
\usepackage{graphicx}
\usepackage{subfigure}
\usepackage{amsfonts}
\usepackage{multirow}
\usepackage{listings}
\usepackage{url}
\usepackage{CJKutf8}
\usepackage{graphicx}
\usepackage{subfigure}
\usepackage{supertabular}
\usepackage{subfigure}
\usepackage{dcolumn,booktabs}
\usepackage{tikz}
\usepackage{xspace}
\usepackage{authblk}

\title{\textsc{LangSAMP}: Language-Script Aware Multilingual Pretraining}

\author[1,2,*]{\bf Yihong Liu}
\author[1,2,*]{\bf Haotian Ye}
\author[1,2]{\bf Chunlan Ma}
\author[1,2,3]{\bf Mingyang Wang}
\author[1,2]{\bf Hinrich Sch\"utze}

\affil[1]{Center for Information and Language Processing, LMU Munich} \affil[2]{Munich Center for Machine Learning (MCML)} \affil[3]{Bosch Center for Artificial Intelligence
 \protect\\ \texttt{\{yihong, yehao, chunlan, mingyang\}@cis.lmu.de}}

\def\secref#1{\S\ref{sec:#1}}
\def\seclabel#1{\label{sec:#1}}

\newcounter{notecounter}
\newcommand{\enotesoff}{\long\gdef\enote##1##2{}}
\newcommand{\enoteson}{\long\gdef\enote##1##2{{
\stepcounter{notecounter}
{\large\bf
\hspace{1cm}\arabic{notecounter} $<<<$ ##1: ##2
$>>>$\hspace{1cm}}}}}
\enoteson
\enotesoff

\begin{document}

\def\frameworkname{\textsc{LangSAMP}\xspace}
\maketitle

\def\thefootnote{*}\footnotetext{Equal contribution.}\def\thefootnote{\arabic{footnote}}

\begin{abstract}

Recent multilingual pretrained language models (mPLMs) often avoid using language embeddings -- learnable vectors assigned to individual languages. 
However, this places a significant burden on token representations to encode all language-specific information, which may hinder language neutrality.
To address this limitation, we propose \textbf{Lang}uage-\textbf{S}cript \textbf{A}ware \textbf{M}ultilingual \textbf{P}retraining (\textbf{\textsc{LangSAMP}}), a method that incorporates both \textbf{language} and \textbf{script} embeddings to enhance representation learning.
Specifically, we integrate these embeddings into the output of the Transformer blocks before passing the final representations to the language modeling head for prediction.
We apply \textsc{LangSAMP} to the continual pretraining of XLM-R \citep{conneau-etal-2020-unsupervised} on a highly multilingual corpus covering more than 500 languages.
The resulting model consistently outperforms the baseline in zero-shot crosslingual transfer across diverse downstream tasks.
Extensive analysis reveals that language and script embeddings capture language- and script-specific nuances, which benefits more language-neutral representations, proven by improved pairwise cosine similarity. In our case study, we also show that language and script embeddings can be used to select better source languages for crosslingual transfer.
We make our code and models publicly available at
\url{https://github.com/cisnlp/LangSAMP}.

\end{abstract}

\section{Introduction}
Encoder-only mPLMs are often regarded as universal text encoders \citep{cer2018universal,huang-etal-2019-unicoder,yang-etal-2020-multilingual}, where the sentence-level or token-level representations are applied to various downstream tasks across different languages \citep{wei2021universal}. 
One of the most attractive aspects of these representations
is their utility in crosslingual
transfer \citep{zoph-etal-2016-transfer,wu-dredze-2019-beto,artetxe-etal-2020-cross}.
That is, representations from a single source
language can be used to fine-tune a multilingual task-specific model (e.g., an mPLM + a task-specific classifier). The fine-tuned model can be applied directly to other languages, without further training. Such a pipeline is particularly useful for low-resource languages, where training data is often scarce \citep{artetxe-etal-2020-call}.

\begin{figure}
    \centering
    \includegraphics[width=0.48\textwidth]{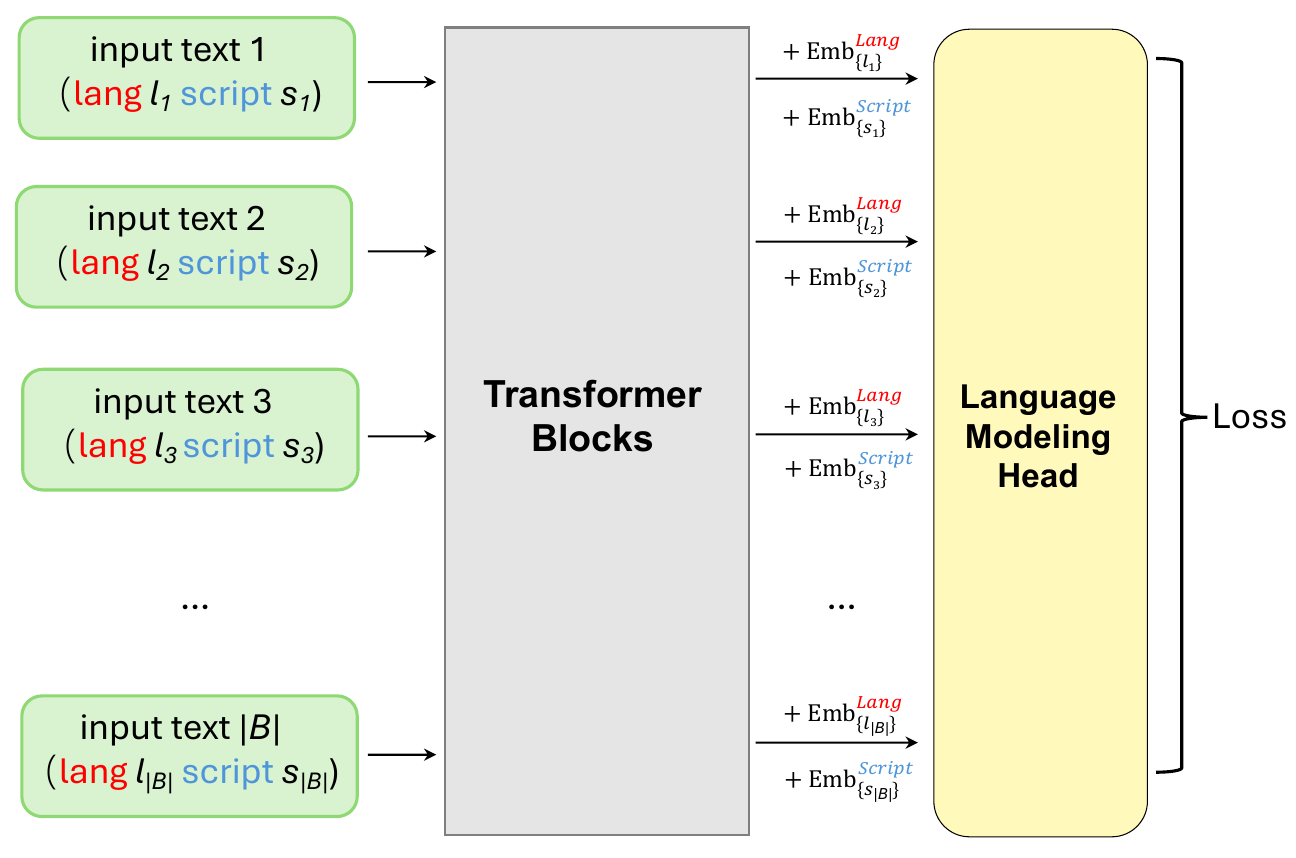}
    \caption{An illustration of \frameworkname for a single batch. Each text may come from different languages and different scripts. Language and script embeddings are added to the transformer output before feeding into the language modeling head. This setup improves the language neutrality of the representations as the auxiliary embeddings share the burden by encoding some language- and script-specific information useful for decoding specific tokens in masked language modeling.}
    \label{fig:first_page_figure}
\end{figure}

The effectiveness of this pipeline depends on the transferability of crosslingual representations. However, previous studies have shown that the representations from recent mPLMs encode a lot of language- and script-specific information \citep{9053443,chang-etal-2022-geometry,wen-yi-mimno-2023-hyperpolyglot}. This is generally not advantageous, as language neutrality, i.e., representations from different languages share a unified subspace, is important for effective crosslingual transfer \citep{libovicky-etal-2020-language,chang-etal-2022-geometry,hua-etal-2024-mothello}. While some approaches attempt to post-align these representations \citep{Cao2020multilingual,pan-etal-2021-multilingual,liu2024translico,xhelili2024breaking}, limited efforts have focused on enhancing language neutrality from the architectural perspective of mPLMs during pretraining.

Early mPLMs, such as XLM \citep{NEURIPS2019_c04c19c2}, leverage language embeddings -- learnable vectors assigned to different languages. These embeddings are added to the token embeddings before being fed into the transformer \citep{vaswani2017transformer} blocks, aiming to alleviate the burden of encoding language-specific information within the token embeddings.
Language embeddings can also guide generation toward the correct target language in machine translation \citep{NEURIPS2019_c04c19c2,song2019mass,liu-etal-2022-flow}. However, more recent mPLMs, such as XLM-R \citep{conneau-etal-2020-unsupervised} and mBERT \citep{devlin-etal-2019-bert}, have discarded these embeddings. The two primary reasons are that (1) mPLMs are expected to have a single, unified parameter set for all languages, and (2) they need to function seamlessly as universal text encoders without requiring language IDs as input. 
However, the removal inevitably reduces the language neutrality of token embeddings and representations (contextual token embeddings), which may negatively impact crosslingual transfer.

To address this limitation, this work proposes \textbf{Lang}uage-\textbf{S}cript \textbf{A}ware \textbf{M}ultilingual \textbf{P}retraining (\textbf{\frameworkname}), a method that incorporates both \textbf{language} and \textbf{script} embeddings to facilitate better representation learning. 
Instead of adding these embeddings to the token embeddings before feeding them into the transformer blocks, we add them to the output of the transformer blocks (final contextual token embeddings) \textbf{before feeding them into the language modeling head}, as shown in Figure \ref{fig:first_page_figure}. 
In the pretraining phase, language and script IDs are required to obtain language and script embeddings, offloading the burden and helping decode specific tokens in masked language modeling.
After pretraining, the backbone (token embeddings and transformer blocks) can function seamlessly as a universal text encoder, which can be fine-tuned together with a task-specific classifier for downstream tasks, without any language or script IDs as input, which are the same as most recent mPLMs.

To validate our approach, we continually pretrain XLM-R \citep{conneau-etal-2020-unsupervised} using \frameworkname on Glot500-c \citep{imanigooghari-etal-2023-glot500}, a multilingual dataset containing over 500 languages. We evaluate the resulting model across a diverse set of downstream tasks, including sentence retrieval, text classification, and sequence labeling, consistently achieving superior performance compared to the baseline. 
We show that better language neutrality is achieved -- \frameworkname improves the pairwise cosine similarity across languages.
Additionally, we observe that language and script embeddings encapsulate typological features, making their similarities a useful resource for selecting optimal source languages in crosslingual transfer.

Our main contributions are as follows: (i) We propose \frameworkname, an effective multilingual pretraining method to improve the language neutrality of representations. 
(ii) We conduct extensive experiments across a spectrum of downstream tasks, demonstrating that our method consistently improves crosslingual transfer performance.
(iii) Our case study shows that language embeddings, as a byproduct, can effectively assist in selecting the optimal source language for crosslingual transfer.

\section{Related Work}

\subsection{Multilingual Pretrained Language Models}

Multilingual pretrained language models (mPLMs) are models that are trained on many languages, with one or multiple self-supervised objectives, such as masked language modeling (MLM) \citep{devlin-etal-2019-bert} or causal language modeling \citep{radford2019language}. These models can be generally classified as encoder-only \citep{devlin-etal-2019-bert,conneau-etal-2020-unsupervised,liang-etal-2023-xlm}, encoder-decoder \citep{liu-etal-2020-multilingual-denoising,fan2021beyond,xue-etal-2021-mt5}, and decoder-only models \citep{lin-etal-2022-shot,shliazhko2022mgpt,scao2022bloom}.
Decoder-only models that have considerably many parameters and are pretrained on a lot of data are also referred to as large language models (LLMs) \citep{achiam2023gpt,touvron2023llama,ustun2024aya}, which are good at natural language generation tasks, typically for high- and medium-resource languages. In parallel, some recent encoder-only models attempt to scale \textit{horizontally}, i.e., cover more languages, especially low-resource ones \citep{ogueji-etal-2021-small,alabi-etal-2022-adapting,imanigooghari-etal-2023-glot500,liu-etal-2024-ofa}. These highly multilingual encoder-only models are particularly good at understanding tasks in a zero-shot crosslingual fashion.

\subsection{Language Embeddings}
Language embeddings are vectors that explicitly or implicitly capture the linguistic characteristics of languages. Early works construct such embeddings using prior knowledge of the languages, resulting in vectors where each dimension encodes a specific linguistic feature \citep{ostling-2015-word,ammar-etal-2016-many,littell-etal-2017-uriel}. However, such features have to be manually defined and may be unavailable for less-studied languages \citep{yu-etal-2021-language}. Therefore, researchers also explore learning language embeddings directly from parallel corpora \citep{malaviya-etal-2017-learning,ostling-tiedemann-2017-continuous,bjerva-augenstein-2018-phonology,tan-etal-2019-multilingual,liu-etal-2023-crosslingual,chen-etal-2023-colex2lang} or monolingual corpora \citep{NEURIPS2019_c04c19c2,yu-etal-2021-language}. This is usually done by assigning an ID to each language, initializing a fixed-length learnable vector, and integrating the vector into the input from that language. The embeddings can capture linguistic features and help crosslingual tasks, e.g., guiding language-specific generation in machine translation in XLM \cite{NEURIPS2019_c04c19c2}. This line of approaches requires language IDs as input for both pretraining and downstream fine-tuning. In contrast, language embeddings are only leveraged in our pretraining. The backbone can be used as a universal text encoder without language IDs for fine-tuning on downstream tasks.

\section{Methodology}
We present \textbf{\frameworkname}, an approach that incorporates both \textbf{language} and \textbf{script} embeddings to facilitate learning more language-neutral representations in multilingual pretraining. \frameworkname preserves the same architecture as the most recent multilingual encoder-only models, except for requiring auxiliary language and script IDs/embeddings in pretraining. In the fine-tuning stage, these auxiliary IDs and embeddings are not required. We introduce the key components in the following.

\begin{figure*}
    \centering
    \includegraphics[width=0.80\textwidth]{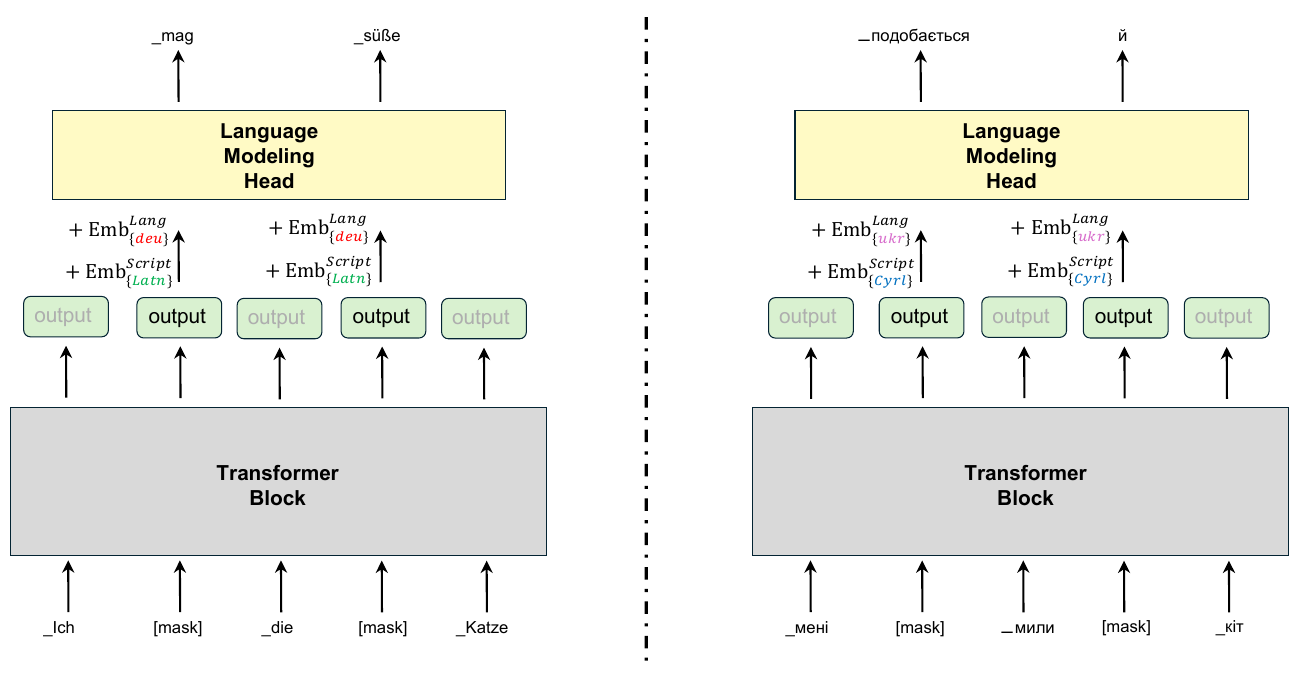}
    \caption{Illustration of \frameworkname applied to a German sentence (left) and a Ukrainian sentence (right), both meaning ``I like the cute cat". Language and script embeddings are added to the outputs from the transformer block. The resulting representation is used to predict the original tokens at the [mask] positions in MLM training.}
    \label{fig:framework}
\end{figure*}

\subsection{Language and Script Embeddings}

Language and script embeddings are introduced to share the
token representations' burden of encoding language- and
script-specific information.  Let
$\boldsymbol{E}^{Lang} \in \mathbb{R}^{L \times D}$ and
$\boldsymbol{E}^{Script} \in \mathbb{R}^{S \times D}$ be the
language and script embeddings, respectively, where $L$ is
the number of languages, $S$ is the number of scripts, and
$D$ is the embedding dimension of the model.  We use
$\boldsymbol{E}^{Lang}_{l}$
(resp. $\boldsymbol{E}^{Script}_{s}$) to denote the
embedding of a specific language $l$ (resp. script
$s$). Similar to token embeddings (which represent  relations between
tokens in vector space), the
language/script embeddings are also expected to capture
structural and typological similarities of languages
(\secref{visualization}) and be useful for selecting good
source language for crosslingual transfer
(\secref{source_language}).

\subsection{Language-Script Aware Modeling}

In the standard MLM pretraining, Transformer blocks generate the final representation at a masked position. 
Subsequently, this representation is fed to the language modeling head to reconstruct the original token. 
Since the original token is used by a specific language and written in a specific script, language- or script-specific information is particularly necessary to decode this token.
From this perspective, the Transformer output used for decoding is not language-neutral by nature.
Our intuition is that we can ease the decoding by giving hints (e.g., the token should be generated in a specific language or script) to the language modeling head. 
In this way, the output of the Transformer blocks does not need
to encode much language- and script-specific information,
and can thus be more language-neutral.
Inspired by this, we add language and script embeddings to the output of Transformer blocks and feed the resulting representations to the language modeling head for decoding, as shown in Figure \ref{fig:framework}. 

Formally, let a training instance (an input sentence) be $X
= [x_1, x_2, \cdots, x_n]$ that comes from language $l$ and
is written in script $s$. We feed $X$ into Transformer blocks and obtain the final contextualized embeddings from the last layer: $\boldsymbol{H} = [\boldsymbol{h}_1, \boldsymbol{h}_2, \cdots, \boldsymbol{h}_n]$. 
We then add the language and script embedding to these outputs to form the final representations: $\boldsymbol{o}_i = \boldsymbol{h_i} + \boldsymbol{E}^{Lang}_{l} + \boldsymbol{E}^{Script}_{s}$. 
The final representations at the masked positions are used to decode the original tokens in MLM:
\begin{equation*}
    \mathcal{L}_{MLM} = - \sum_{i \in \mathcal{M}} \log P_{MLM}(x_i | \boldsymbol{o_i})
\end{equation*}
where $\mathcal{M}$ is the set of masked positions in $X$ and $P_{MLM}(x_i | \boldsymbol{o_i})$ is the probability of decoding the original token $x_i$ given the final representation $\boldsymbol{o}_i$, which is computed by the language modeling head. Since $\boldsymbol{E}^{Lang}_{l}$ and $\boldsymbol{E}^{Script}_{s}$ provide language and script-specific information, we expect that $\boldsymbol{h}_i$ will be more language-neutral (\secref{Similarity}), which is beneficial to zero-shot crosslingual transfer (\secref{results}).

\subsection{Fine-tuning on Downstream tasks}
Since we only leverage language and script embedding in the pretraining for MLM, the core architecture (token embeddings + Transformer blocks) remains the same as most mainstream mPLMs, such as XLM-R. 
In this way, we \textbf{do not} need any language or script IDs as input to obtain the Transformer output, i.e., the final contextualized embeddings $\boldsymbol{H}$. 
This means our pretrained model can be fine-tuned in the
standard way in the NLP pipeline. 
Specifically, for any downstream tasks that require a task-specific classifier (either token-level or sequence-level tasks), we can feed the final contextualized embeddings $\boldsymbol{H} = [\boldsymbol{h}_1, \boldsymbol{h}_2, \cdots, \boldsymbol{h}_n]$ to the classifier and update the model parameters according to the fine-tuning objective, where language or script embeddings are not participating at all. 
In addition, as $\boldsymbol{H}$ is more language-neutral thanks to \frameworkname, we expect the representations to boost zero-shot crosslingual transfer (\secref{results}).

It is important to note that we do not increase the number of parameters used to compute token or sentence representations, as the auxiliary language/script embeddings are employed only during pretraining.
This contrasts with prior work, which often introduces additional components -- such as Adapters -- during downstream fine-tuning \citep{pfeiffer-etal-2022-lifting,balne2024parameterefficientfinetuning}. 
Consequently, any improvements in downstream task performance can only be attributed to enhanced representations learned by the Transformer itself, rather than to added model capacity because of more parameters.

\section{Experiments}

\subsection{Setups}\seclabel{setups}

\paragraph{Training Corpora and Tokenizer} We use Glot500-c \citep{imanigooghari-etal-2023-glot500}, a corpus that has monolingual data from more than 500 languages written in 30 different scripts. We treat each language-script as a separate entity and refer to those covered by XLM-R \citep{conneau-etal-2020-unsupervised} as \emph{head languages}, whereas the remaining are \emph{tail languages} (also low-resource languages). We use the tokenizer of Glot500-m \citep{imanigooghari-etal-2023-glot500}, which is a SentencePiece Unigram tokenizer \citep{kudo-richardson-2018-sentencepiece, kudo-2018-subword} whose vocabulary is merged from the subwords in XLM-R and new subwords learned from Glot500-c.

\begin{table*}[ht]
    \scriptsize
    \footnotesize
    \setlength{\tabcolsep}{0.5mm}{}
    \resizebox{\textwidth}{!}{%
    \begin{tabular}{lcc|cc|cc|cc|cc}
        \toprule
        & \multicolumn{2}{c}{tail} & \multicolumn{2}{c}{head} & \multicolumn{2}{c}{Latn} & \multicolumn{2}{c}{non-Latn} & \multicolumn{2}{c}{all}  \\ 
        \cmidrule(rr){2-3} \cmidrule(rr){4-5} \cmidrule(rr){6-7} \cmidrule(rr){8-9} \cmidrule(rr){10-11}
        & Baseline & \frameworkname & Baseline & \frameworkname & Baseline & \frameworkname & Baseline & \frameworkname & Baseline & \frameworkname \\
        \midrule
SR-B	&	36.9 (0.0)  & 	\textbf{39.5} (0.0)  & 	60.6 (0.0)  & 	\textbf{61.3} (0.0)  & 	40.7 (0.0)  & 	\textbf{42.8} (0.0)  & 	51.2 (0.0)  & 	\textbf{53.5} (0.0)  & 	42.9 (0.0)  & 	\textbf{45.1} (0.0)  \\
SR-T	&	56.9 (0.0)  & 	\textbf{58.6} (0.0)  & 	74.8 (0.0)  & 	\textbf{76.1} (0.0)  & 	67.5 (0.0)  & 	\textbf{68.7} (0.0)  & 	73.7 (0.0)  & 	\textbf{75.6} (0.0)  & 	69.7 (0.0)  & 	\textbf{71.1} (0.0)  \\
Taxi1500	&	47.1 (4.8)  & 	\textbf{50.8} (2.4)  & 	59.9 (2.9)  & 	\textbf{61.2} (1.2)  & 	48.2 (4.6)  & 	\textbf{51.7} (2.1)  & 	58.8 (3.1)  & 	\textbf{60.1} (1.7)  & 	50.3 (4.2)  & 	\textbf{53.4} (2.0)  \\
SIB200	&	69.0 (1.4)  & 	\textbf{70.2} (1.9)  & 	82.2 (1.4)  & 	\textbf{82.6} (1.2)  & 	72.1 (1.3)  & 	\textbf{73.1} (1.8)  & 	81.1 (1.5)  & 	\textbf{81.7} (1.2)  & 	75.0 (1.3)  & 	\textbf{75.9} (1.6)  \\
NER	&	60.1 (0.6)  & 	\textbf{60.8} (0.8)  & 	64.0 (0.6)  & 	\textbf{64.1} (0.6)  & 	67.0 (0.5)  & 	\textbf{67.6} (0.6)  & 	\textbf{53.9} (0.7)  & 	\textbf{53.9} (0.5)  & 	62.2 (0.5)  & 	\textbf{62.6} (0.6)  \\
POS	&	61.3 (1.0)  & 	\textbf{61.4} (0.9)  & 	76.0 (0.4)  & 	\textbf{76.2} (0.4)  & 	\textbf{74.6} (0.5)  & 	74.5 (0.4)  & 	66.2 (1.0)  & 	\textbf{66.8} (0.8)  & 	71.5 (0.6)  & 	\textbf{71.6} (0.5)  \\
        \bottomrule
    \end{tabular}
    }
    \caption{Performance of \frameworkname and baseline on six downstream tasks across five random seeds. We report the performance by grouping languages according to two characteristics: (1) whether it is a head or a tail language, and (2) whether it is written in Latin script or non-Latin script. The average performance within each group and the standard deviation (in parentheses) are computed. \frameworkname consistently achieves on-par performance or outperforms the baseline across all groups and downstream tasks. \textbf{Bold}: best result for each group in each task.}
    \label{tab:main_perf}
\end{table*}

\paragraph{Continued pretraining} We use the weights from XLM-R to initialize our \frameworkname model for MLM pretraining. \textbf{Language and script embeddings are randomly initialized with dimensions $\mathbb{R}^{610 \times 768}$ and $\mathbb{R}^{30 \times 768}$ respectively}. We continually train our model on Glot500-c, where we sample data from a multinomial distribution with a temperature of 0.3, to increase the amount
of training instances of low- and medium-resource languages.
We use AdamW optimizer \citep{ba2015adam,loshchilov2018decoupled} with $(\beta_1, \beta_2) = (0.9, 0.999)$ and $\epsilon = \text{1e-6}$. 
The initial learning rate is set to 5e-5. The effective batch size is 1,024 in each training step, where the gradient accumulation is 8 and the per-GPU batch size is 32. 
We train the model on 4 NVIDIA RTX6000 GPUs.
Each training instance in a batch contains sentences from \textbf{the same language-script} which are concatenated to a chunk of 512 tokens.
Each batch contains instances from \textbf{different language-scripts}.
We store checkpoints every 5K steps and apply early stopping with the best average performance on downstream tasks. We set the maximum steps to 150K. 
The training takes about 4 weeks. 

\paragraph{Baseline} To validate \frameworkname, we create a baseline where language and script embeddings are not used. 
This baseline can be regarded as a reproduction of Glot500-m \citep{imanigooghari-etal-2023-glot500}. 
For a fair comparison, the training hyperparameters and training data (100\% data of Glot500-c) are the same as \frameworkname. 
However, in our ablation study \secref{ablation}, due to a constrained computing budget, we cannot continually pretrain model variants on full Glot500-c for validating each component individually (with/without language or script embeddings). 
Instead, we create such variants and pretrain them using a small portion (5\%) of Glot500-c.
As a result, the baseline model in Table \ref{tab:main_perf} is different from the vanilla model in Table \ref{tab:ablation}.

\subsection{Downstream Tasks}

We consider the following three evaluation types, with two datasets for each type. The evaluation is
done in an English-centric zero-shot crosslingual transfer style for evaluation types that require fine-tuning. 
That is, we first fine-tune the pretrained model on the English train set, then select the
best checkpoint on the English development set, and finally evaluate the best checkpoint on the test sets of all other languages.
For Sentence Retrieval, which does not involve any fine-tuning, we simply use English as the retrieval query language. 
For all tasks, only a subset of languages (head and tail languages) supported by Glot500-c are considered. We show the detailed information of the used dataset and hyperparameter settings in \secref{hyperparam}. We introduce the evaluation types and datasets in the following.

\paragraph{Sentence Retrieval.} We use Bible (SR-B) and Tatoeba \citep{artetxe-schwenk-2019-massively} (SR-T). The pairwise similarity for retrieving the target sentences is calculated using the mean pooling of contextualized word embeddings at the 8th layer. 

\paragraph{Text Classification.} We use Taxi1500 \citep{ma2023taxi1500} and SIB200 \citep{adelani-etal-2024-sib}. The former is a dataset based on the Bible, whereas the latter is based on FLORES-200 \citep{costa2022no} with more modern genres like technology.

\paragraph{Sequence Labeling.} We use WikiANN for named entity recognition (NER) \citep{pan-etal-2017-cross} and Universal Dependencies \citep{de-marneffe-etal-2021-universal} for Part-Of-Speech (POS) tagging.

\subsection{Results and Discussion}\seclabel{results}
We evaluate the \frameworkname model and baseline to understand how the integration of language and script embeddings influences crosslingual transfer. 
We group the transfer target languages based on two characteristics: (1) whether it is a head or tail language, and (2) whether it is written in Latin or a non-Latin script. This grouping aims to directly identify the effectiveness of \frameworkname on low-resource languages and languages written in a less common script. The results are shown in Table \ref{tab:main_perf}.

\begin{table*}[ht]
    \footnotesize
    \centering
    \setlength{\tabcolsep}{0.66mm}{}
    \begin{tabular}{lcccccccccccccccccc}
        \toprule
        & \multicolumn{3}{c}{SR-B} & \multicolumn{3}{c}{SR-T} & \multicolumn{3}{c}{Taxi1500} &  \multicolumn{3}{c}{SIB200} & \multicolumn{3}{c}{NER} & \multicolumn{3}{c}{POS}\\
        \cmidrule(lr){2-4} \cmidrule(lr){5-7} \cmidrule(lr){8-10} \cmidrule(lr){11-13} \cmidrule(lr){14-16} \cmidrule(lr){17-19}
        & tail & head & all & tail & head & all & tail & head & all & tail & head & all & tail & head & all & tail & head & all\\
        \midrule
vanilla model & 11.9 & 56.4 & 23.2 & 46.0 & 77.7 & 68.6 & 18.1 & \underline{58.6} & 28.4 & 56.1 & \textbf{83.0} & 68.3 & \underline{55.1} & \underline{62.8} & \underline{59.3} & \underline{49.9} & 75.7 & \underline{67.8} \\
w/ $\boldsymbol{E}^{Lang}$ & \underline{13.1} & \underline{57.9} & \underline{24.5} & \textbf{49.1} & \underline{79.0} & \underline{70.5} & 18.3 & 58.5 & \underline{28.5} & \underline{57.2} & \underline{82.7} & \underline{68.8} & \textbf{55.2} & \textbf{63.0} & \textbf{59.5} & \underline{49.9} & \underline{75.8} & \underline{67.8} \\
w/ $\boldsymbol{E}^{Script}$ & 12.5 & 57.4 & 23.9 & \underline{48.3} & 78.4 & 69.8 & \underline{18.5} & 57.0 & 28.2 & 56.6 & 82.1 & 68.2 & \underline{55.1} & 62.4 & 59.0 & \textbf{50.8} & \textbf{76.2} & \textbf{68.4} \\
w/ $\boldsymbol{E}^{Lang}$ and $\boldsymbol{E}^{Script}$ & \textbf{13.4} & \textbf{58.7} & \textbf{24.9} & \textbf{49.1} & \textbf{79.5} & \textbf{70.8} & \textbf{20.6} & \textbf{58.8} & \textbf{30.3} & \textbf{57.9} & \textbf{83.0} & \textbf{69.3} & 54.9 & 61.6 & 58.6 & 49.7 & 75.6 & 67.6 \\
 \bottomrule
    \end{tabular}
    \caption{Ablation study. We investigate the effectiveness of language and script embeddings on downstream performance. Note that the vanilla model and $\text{w} /\boldsymbol{E}^{Lang}$ and $\boldsymbol{E}^{Script}$ are different from Baseline and \frameworkname in Table \ref{tab:main_perf} because of the smaller pretraining data size. By including both types of embeddings, the model achieves the overall best performance among all variants. \textbf{Bold} (\underline{underlined}): best (second-best) result for each column.}
    \label{tab:ablation}
\end{table*}

\paragraph{Both tail and head languages benefit.} 
We observe consistent improvements in tail and head languages across tasks. The enhancement is more obvious in tail languages. For example, \frameworkname improves the performance by 7\% for tail languages vs 1\% for head languages in SR-B. A similar phenomenon can also be seen for other tasks. This pattern indicates that \frameworkname can be more helpful for those tail languages, for which the training data is scarce. With the help of language embeddings sharing the burden, the \frameworkname model can have more language-neutral representations for these languages, resulting in better performance.

\paragraph{Both non-Latin and Latin languages benefit.} 
We observe similar consistent improvements when grouping languages into Latin or non-Latin languages. Different from the trend seen in tail/head groups, we see that no group shows an obvious larger enhancement compared to the other group. This can be explained by the fact that head and tail languages are distributed more equally in Latn and non-Latn groups. In addition, the improvements indicate the incorporation of script embeddings is helpful. By decoupling some script-specific information from the representations, the output generated by the backbone is more script-neutral, leading to better crosslingual transfer across scripts.

\paragraph{Improvements can vary slightly across tasks.} We
    observe more consistent large improvements for
    sequence-level tasks -- retrieval and classification --
    where \frameworkname outperforms the baseline in all
    groups. However, on sequence labeling
    tasks, \frameworkname achieves very close performance to
    the baseline. For example, \frameworkname scores are 0.1 less compared to the baseline on NER. This could be related to the difficulty of the tasks: both NER and POS are relatively easy tasks, and models
can transfer well in prevalent classes, e.g., \textit{nouns}, through shared vocabulary \citep{imanigooghari-etal-2023-glot500,liu-etal-2024-ofa}. Therefore, decoupling language- or script-specific information from the Transformer output can be less helpful for these tasks. Nevertheless, the overall improvements across tasks indicate the superiority of \frameworkname compared with the baseline.

\section{Analysis}

\subsection{Ablation Study}\seclabel{ablation}
In the ablation study, we want to explore the effectiveness
of language embeddings and script
embeddings individually. However, due to a limited computation budget, we cannot run experiments on the full corpora for each variant. Therefore, we select 5\% data for each language from Glot500-c and continually pretrain XLM-R using the same hyperparameters used in the main experiments described in \secref{setups}. Specifically, we consider four variants: \textbf{a}) model without language/script embeddings; \textbf{b}) model with only language embeddings; \textbf{c}) model with only script embeddings; and \textbf{d}) model with both language and script embeddings. The performance of each variant is shown in Table \ref{tab:ablation}.

\paragraph{Either language or script embeddings help.} 
The vanilla model achieves the overall worst performance among all model variants.
As long as language or script embeddings are included, we generally observe a consistent improvement across all downstream tasks.
This indicates that both language and script embeddings can share the burden of encoding too much language- and script-specific information in the token representations.
As a result, the representations generated by the model variants with language or script embeddings are more language-neutral, benefiting the crosslingual transfer. 
The best overall performance is achieved when both language and script embeddings are used, suggesting that decoupling both language- and script-specific information would be the best option for improving crosslingual transfer.

\begin{figure*}
    \centering
    \includegraphics[width=0.45\textwidth]{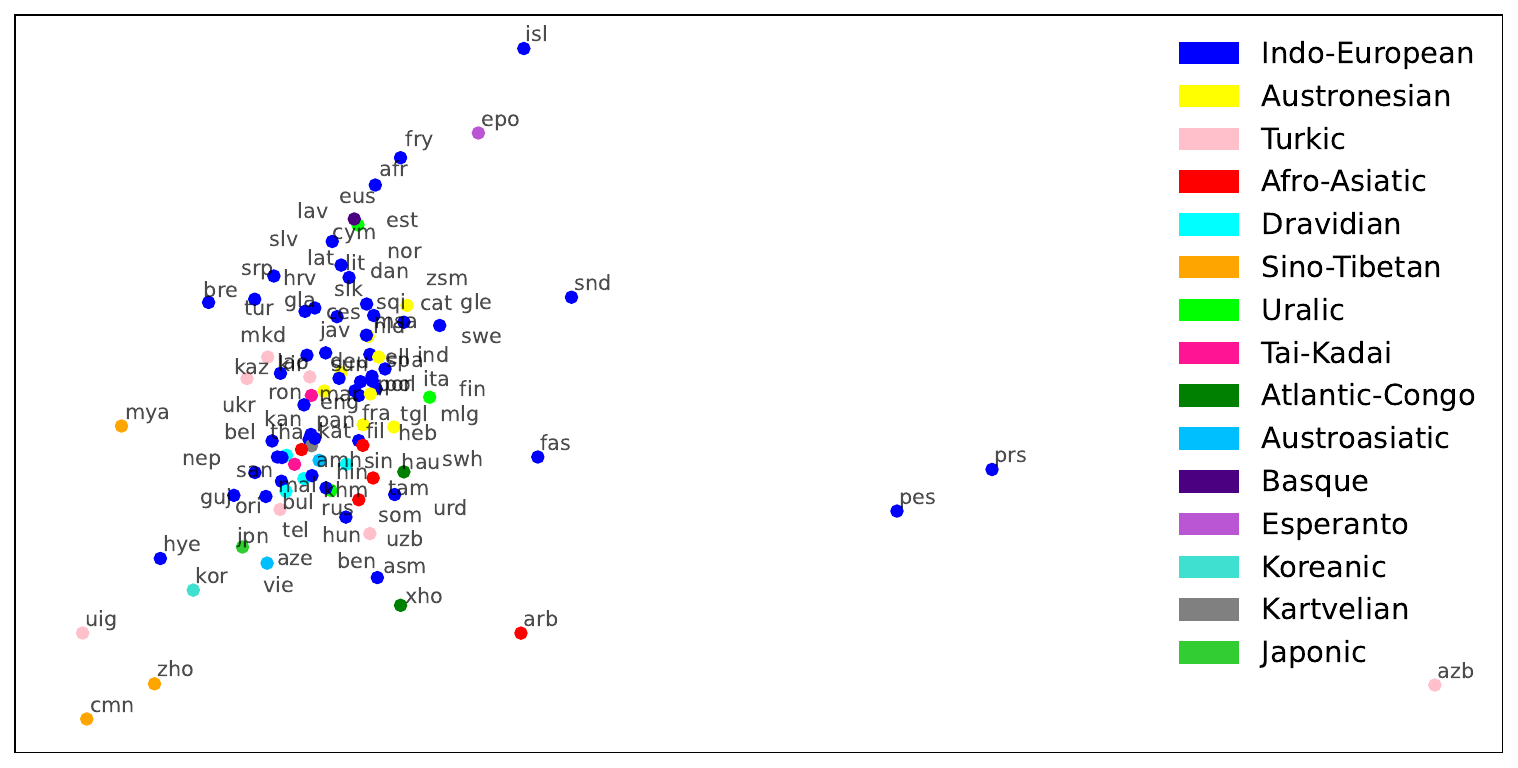}
    \includegraphics[width=0.45\textwidth]{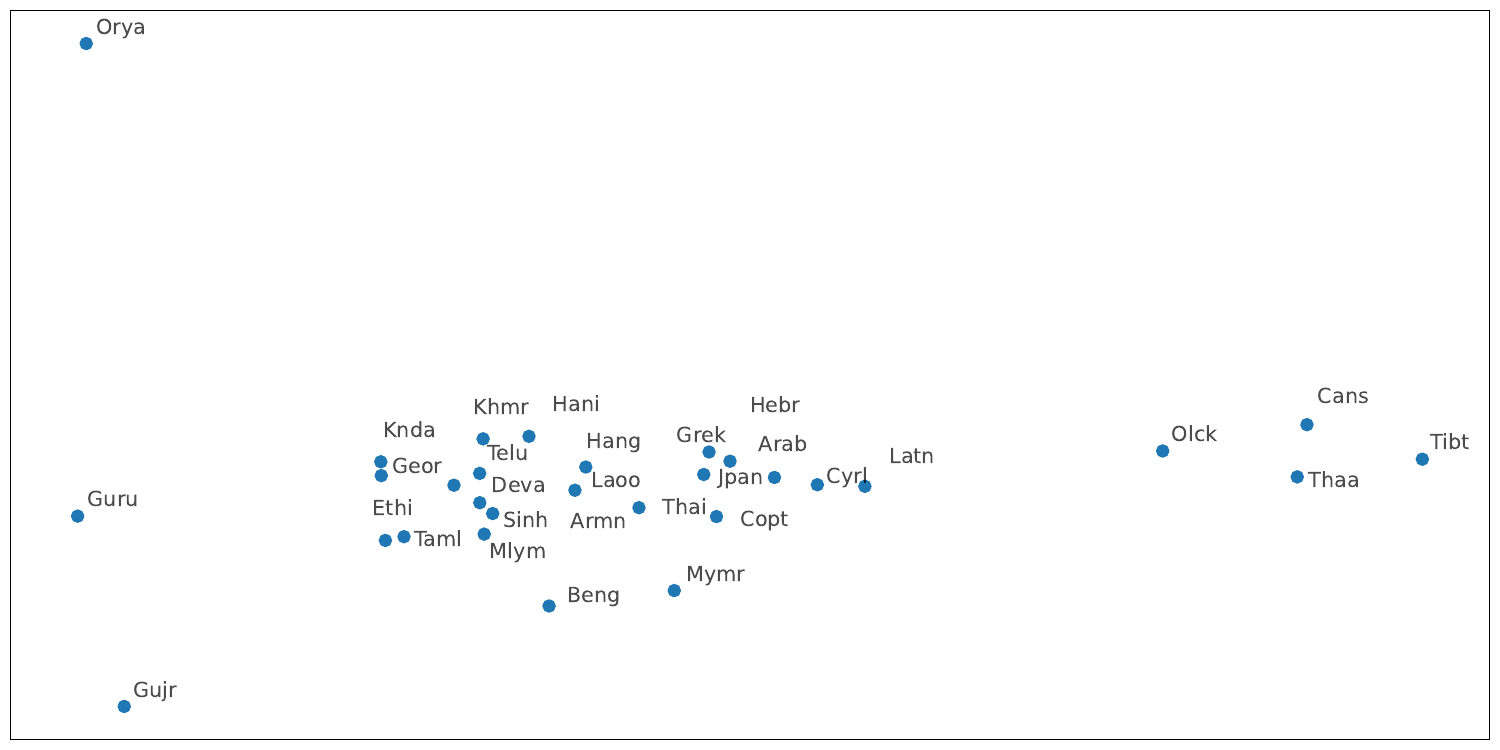}
    \caption{PCA visualizations of head language embeddings (left) and script embeddings (right).
    We see that some related languages and scripts are close to each other, indicating that they encode language- and script-specific information. 
    Data imbalance may have caused some languages/scripts with limited data to appear as outliers.}
    \label{fig:pca_emb}
\end{figure*}

\paragraph{Improvement varies across task types.} 
Similar to the findings in \secref{results}, we observe that including the auxiliary embeddings is very helpful for sequence-level tasks, especially sentence retrieval, where we observe the highest enhancement, while less helpful for token-level tasks. 
It is also noticeable that including language embeddings is the most effective for sentence retrieval (either best or the second best per column).
On the other hand, the sequence labeling task does not enjoy large improvements: most model variants achieve on-par performance with each other.
The reason has been discussed in \secref{results}: NER and POS are relatively simple tasks since models can transfer easily in prevalent classes.
Nevertheless, the overall results show the effectiveness of the auxiliary embeddings.

\subsection{Qualitative Exploration: Visualization}\seclabel{visualization}

We visualize language and script embeddings in Figure \ref{fig:pca_emb}. Only head language embeddings are chosen for better readability.
We observe that similar or related languages are located close to each other.
For example, \textbf{cmn} and \textbf{zho} (simplified and traditional Chinese, lower left) are closest to each other, as are \textbf{pes} (Iranian Persian) and \textbf{prs} (Dari).
The languages that are mutually influenced by Chinese to a large extent, \textbf{jpn}, \textbf{kor}, and \textbf{vie}, are also close to each other. 
Most European languages, as well as Indian languages that belong to the Indo-European family, form a rather dense cluster in the middle.

In the plot on the right, most scripts of the Indian subcontinent are found close to each other (\textbf{Deva}, \textbf{Telu}, \textbf{Mlym}, \textbf{Taml}, \textbf{Knda}, \textbf{Sinh}, \textbf{Beng}), despite some outliers (e.g., \textbf{Gujr} and \textbf{Guru}), probably due to the small amount of data that is written in these scripts. \textbf{Hani} and scripts of languages that are mutually influenced (\textbf{Hang} and \textbf{Jpan}) are not far from each other.
The same is true for two very related scripts, \textbf{Thai} and \textbf{Laoo}. In summary, the learnable language and script embeddings can capture language- and script-specific information in the training, which can be helpful for the language-neutrality of the output of transformer blocks.

\subsection{Quantitative Exploration: Similarity}\seclabel{Similarity}

\begin{table*}[ht]
    \scriptsize
    \footnotesize
    \centering
    \setlength{\tabcolsep}{1.5mm}{}
    \begin{tabular}{lcc|cc|cc|cc|cc}
        \toprule
        & \multicolumn{2}{c}{tail} & \multicolumn{2}{c}{head} & \multicolumn{2}{c}{Latn} & \multicolumn{2}{c}{non-Latn} & \multicolumn{2}{c}{all}  \\ 
        \cmidrule(rr){2-3} \cmidrule(rr){4-5} \cmidrule(rr){6-7} \cmidrule(rr){8-9} \cmidrule(rr){10-11}
        & English & Donor & English & Donor & English & Donor & English & Donor & English & Donor \\
        \midrule
Taxi1500 & 47.3 & \textbf{48.3} & 59.1 & \textbf{60.3} & 48.4 & \textbf{49.0} & 58.1 & \textbf{60.5} & 50.2 & \textbf{51.2} \\
SIB200	 & \textbf{67.9} & \textbf{67.9} & 81.2 & \textbf{81.6} & 71.0 & \textbf{71.1} & 80.3 & \textbf{80.6} & 74.0 & \textbf{74.2}  \\
NER	     & 61.2 & \textbf{61.7} & 64.1 & \textbf{65.6} & \textbf{67.5} & 66.9 & 54.6 & \textbf{58.5} & 62.8 & \textbf{63.8} \\
POS	     & \textbf{63.2} & 53.8 & \textbf{77.0} & 72.3 & \textbf{75.5} & 68.4 & \textbf{68.1} & 63.6 & \textbf{72.8} & 66.6 \\
        \bottomrule
    \end{tabular}
    \caption{
    Performance of \frameworkname, using English vs
    the closest donor language 
    (based on cosine similarity induced from language
    embeddings) as the source language for zero-shot
    crosslingual transfer.  Each number is the average over
    all target languages in a class.
    \textbf{Bold}: the result that is better for an
    English/Donor comparison.
    }
    \label{tab:donor_perf}
\end{table*}

\begin{figure}
    \centering
    \includegraphics[width=0.46\textwidth]{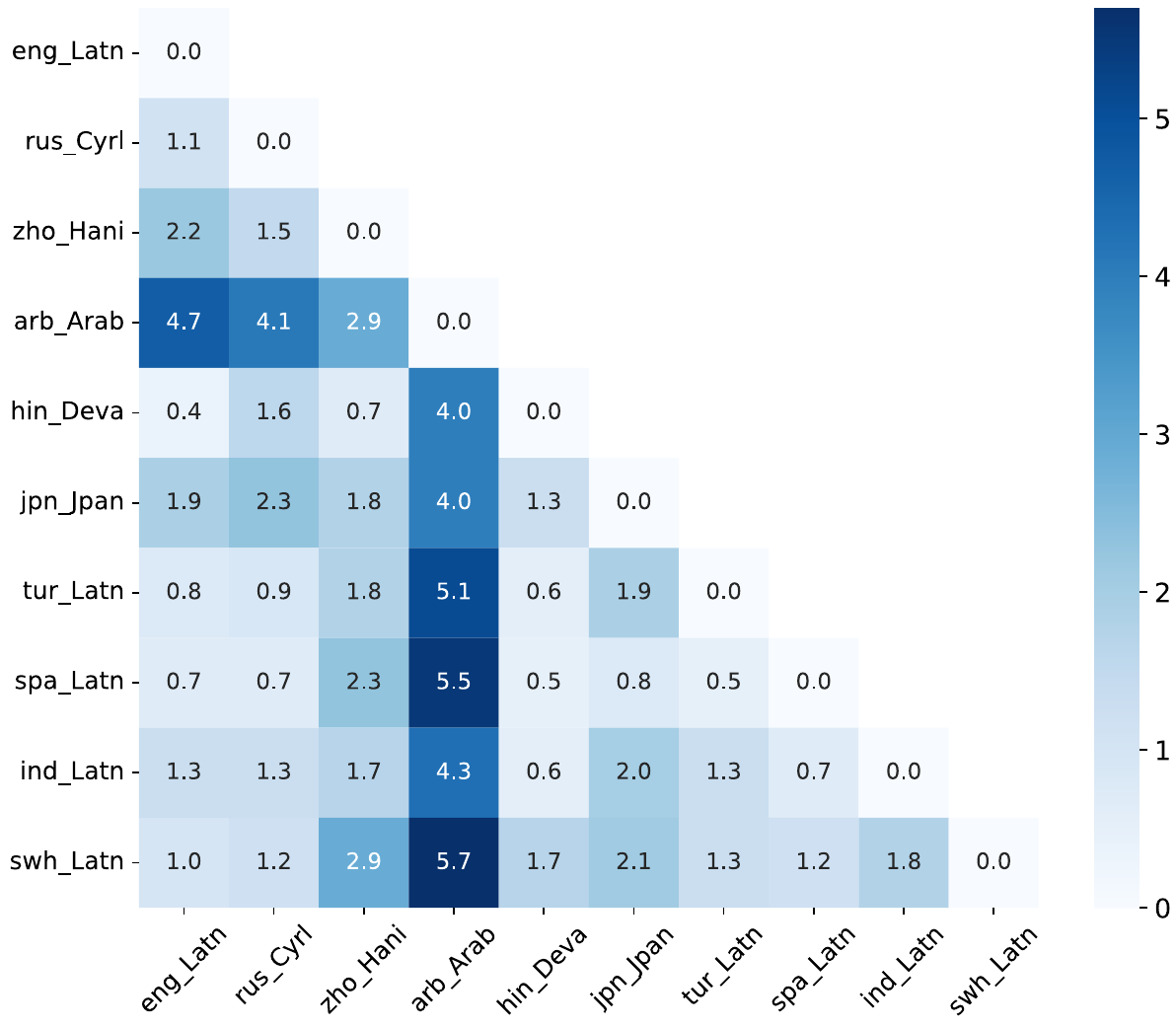}
    \caption{Similarity improvement (by percentage) from baseline to \frameworkname in terms of the pairwise cosine similarity. Similarity is increased for each pair, indicating better language neutrality of the representations.}
    \label{fig:sim_diff}
\end{figure}

We expect that \frameworkname can generate more language-neutral representations, meaning that representations of semantically equivalent sentences from different languages are similar. To evaluate this, we selected 10 high-resource languages that differ typologically and use a diverse set of scripts:
\textbf{eng\_Latn},
\textbf{rus\_Cyrl},
\textbf{zho\_Hani},
\textbf{arb\_Arab},
\textbf{hin\_Deva},
\textbf{jpn\_Jpan},
\textbf{tur\_Latn},
\textbf{spa\_Latn},
\textbf{ind\_Latn},
and
\textbf{swa\_Latn}.
We calculated the pairwise cosine similarity of sentence representations using 100 randomly sampled parallel sentences from SR-B. Sentence representations are obtained by mean-pooling the token representations at the 8th layer, followed by subtracting the language centroid (the average of all 100 sentence representations for that language). We report the pairwise cosine similarity in Figure \ref{fig:sim_comparison} in \secref{appendix:sim} and show the improvement (by percentage) in Figure \ref{fig:sim_diff}.

\begin{table}
    \scriptsize
    \footnotesize
    \centering
    \setlength{\tabcolsep}{1.5mm}{}
    \begin{tabular}{l|cc|cc|cc|cc}
\toprule
\multicolumn{1}{l}{} & \multicolumn{2}{c}{Taxi1500} & \multicolumn{2}{c}{SIB200} & \multicolumn{2}{c}{NER} & \multicolumn{2}{c}{POS} \\
\midrule
\multirow{2}{*}{tha} & eng & jpn & eng & jpn & eng & jpn & eng & jpn \\
 & \textbf{63.8} & \textbf{63.8} & 85.4 & \textbf{85.7} & 2.1 & \textbf{10.2} & \textbf{58.3} & 27.5 \\[0.5em]
\multirow{2}{*}{yue} & eng & zho & eng & zho & eng & zho & eng & zho \\
 & 55.4 & \textbf{67.7} & - & - & 25.7 & \textbf{73.5} & 42.6 & \textbf{80.9} \\[0.5em]
\multirow{2}{*}{san} & eng & hin & eng & hin & eng & hin & eng & hin \\
 & - & - & 72.9 & \textbf{76.6} & 38.4 & \textbf{53.4} & 25.5 & \textbf{32.7} \\[0.5em]
\multirow{2}{*}{urd} & eng & hin & eng & hin & eng & hin & eng & hin \\
 & - & - & 79.1 & \textbf{80.6} & 65.1 & \textbf{76.8} & 69.7 & \textbf{89.7} \\[0.5em]
\multirow{2}{*}{lin} & eng & swh & eng & swh & eng & swh & eng & swh \\
 & 47.1 & \textbf{54.7} & 68.2 & \textbf{73.3} & 47.6 & \textbf{55.9} & - & - \\[0.5em]
\multirow{2}{*}{run} & eng & swh & eng & swh & eng & swh & eng & swh \\
 & 48.0 & \textbf{55.2} & 65.2 & \textbf{72.7} & - & - & - & - \\
\bottomrule
\end{tabular}
    \caption{
    Languages with large improvements when using the closest
    donor language.
    In each task, the first/second column indicates results using English/the donor language as the source language.
    ``-'' indicates the language is not covered by the task.
    \textbf{Bold}: best result for each language in each task.
    }
    \label{tab:donor_examples}
\end{table}

We can observe that the similarity between any two languages is improved in \frameworkname. 
The enhancement is especially noticeable for typologically distinct languages using different scripts.
For example, arb\_Arab is in a different language family and written in a different script compared to the other 9 languages; the similarity involving arb\_Arab is greatly improved: 4.7\% for eng\_Latn and 4.1\% rus\_Cyrl. 
Importantly, since \frameworkname does not incorporate additional parallel data, this improvement is solely attributed to the inclusion of language and script embeddings during pretraining.
This indicates that \frameworkname effectively generates more language-neutral representations by decoupling language- and script-specific features into auxiliary embeddings.

\subsection{Case Study: Source Language Selection}\seclabel{source_language}

Previous studies show language similarities have been useful for selecting good source languages for crosslingual transfer \citep{lin-etal-2019-choosing,lauscher-etal-2020-zero,nie-etal-2023-cross,wang-etal-2023-nlnde, wang-etal-2023-gradsim, lin-etal-2024-mplm}. 
We expect this to also apply to the similarities induced by our language embeddings.
Therefore, we conduct a case study and use the languages mentioned in \secref{Similarity} as the donor languages.
When performing the downstream task for a specific target language, instead of always using English as the source language, we select the donor language that is the most cosine-similar to the target language.
We evaluate the \frameworkname model on Taxi1500, SIB200, NER, and POS in a zero-shot crosslingual transfer style. The aggregated results are reported in Table \ref{tab:donor_perf}, and we select representative target languages that benefit from choosing a good donor language in Table \ref{tab:donor_examples}.

\paragraph{Effects of donor vary across tasks.} 
Our results suggest that the performance gain from using a donor language 
varies across tasks. 
The gain in the text classification task is more consistent than in the sequence labeling task.
We assume the primary reason is that the training data for NER and POS are not parallel, and the size of the training data is highly variable across languages. 
For example, English has much more data than some of the other donor languages for these two tasks.

\paragraph{Non-Latin languages benefit more.} 
For the text classification task, greater improvements can be observed in non-Latin script languages than in Latin script languages. This reflects previous findings that non-Latin script languages are less represented in mPLMs \citep{muller-etal-2021-unseen} and indicates the effectiveness of leveraging language embeddings in selecting better donor languages for them.

\paragraph{Donor is frequently from the same family.} 
We find that language embeddings frequently identify a donor language of the same family as the target language, leading to a large performance improvement over English as the source.
For example, as shown in Table \ref{tab:donor_examples}, \textbf{zho\_Hani} as a donor language for \textbf{yue\_Hani} leads to large performance gains on all three tasks.
Similar gains are seen using \textbf{hin\_Deva} for \textbf{san\_Deva}.
Positive effects can also be found across scripts, as in the case of using \textbf{hin\_Deva} for \textbf{urd\_Arab}, two very similar languages written in different scripts.

\paragraph{Interesting cases of unrelated donors.} 
We also notice some interesting cases where the closest donor language is not or only partially related to the target language, but nevertheless aids transfer performance as shown in Table \ref{tab:donor_examples}. 
For example, \textbf{jpn\_Jpan} has a positive effect for 
\textbf{tha\_Thai}. Similarly, for \textbf{tuk\_Latn}, using \textbf{rus\_Cyrl} as the source achieves better transfer performance than English.

\section{Conclusion}
We propose \frameworkname, a multilingual pretraining approach that leverages auxiliary language and script embeddings to facilitate more language-neutral representations by offloading the burden of encoding language- and script-specific information within the Transformer outputs. 
These embeddings are added to the output of the transformer blocks before being fed into the language modeling head for decoding. 
In this way, we keep the model structure simple, allowing it to function as a universal text encoder, without requiring language or script IDs as input,  while easing the burden of the output encoding too much language- and script-specific information. 
Through extensive experiments, we show \frameworkname consistently outperforms the baseline on various downstream tasks, especially in sequence-level tasks.
Our ablation study confirms the effectiveness of both language and script embeddings. 
\frameworkname exhibits improved language neutrality, as reflected by increased pairwise similarity across all donor languages.
Furthermore, our case study demonstrates that the byproducts -- auxiliary language/script embeddings -- encode language- and script-specific information, which can facilitate the selection of optimal source languages for more effective crosslingual transfer.

\section{Future Direction}

Looking ahead, a promising research direction is to further explore and refine the use of auxiliary language/script embeddings to guide language-neutral representation learning, particularly in the middle layers of multilingual models.
Prior studies have shown that intermediate layers of encoder-only or decoder-only models often exhibit higher language neutrality \citep{jalili-sabet-etal-2020-simalign, imanigooghari-etal-2023-glot500,Zhao2024Multilingualism,li2025exploringmultilingualprobinglarge}.

Future work could investigate ways to explicitly steer middle-layer representations toward greater neutrality, for example, by combining auxiliary embeddings with layer-specific objectives or contrastive learning alignment techniques.

Furthermore, language and script embeddings hold potential for enhancing controlled multilingual generation in decoder-only and encoder-decoder models, enabling more accurate and consistent generation in the target language based on instructions provided in the prompt.

\section*{Limitations}

Due to the constraints of computing resources, we are not able to continue pretraining the model using the full Glot500-c data in \textbf{our ablation study}. However, as all variants are trained in a strictly controlled environment, their results can be compared in a fair way, and the consistent improvement suggests the effectiveness of the language embeddings. 

In addition, we do not consider the possibility of introducing language and script embeddings before the Transformer blocks.
Although this is also a possible architecture, it does not fulfill our aim and therefore is not relevant to us. 
Our primary prerequisite is that the resulting model can work as a universal text encoder without any language or script IDs as input, just like most highly multilingual models (e.g., XLM-R \citep{conneau-etal-2020-unsupervised} and mBERT \citep{devlin-etal-2019-bert}).
\frameworkname only requires language or script IDs in the pretraining stage. 
After that, the backbone (token embeddings + the Transformer blocks) acts exactly as a universal text encoder. 
Investigating whether architectures that integrate language/script embeddings before the Transformer could improve language representations at scale is outside the scope of our work, but we consider it a promising direction for future research.

Another potential limitation is the coverage of languages and scripts. Our model uses 610 languages and 30 scripts from Glot500-c. For low-resource languages not supported by our model, we can still generate representations since language IDs are not required as input. However, without a corresponding language embedding, it becomes challenging to select the optimal donor language for crosslingual transfer. Nonetheless, when adapting to these languages, the language embeddings can be expanded, similar to the approach commonly used for vocabulary extension.

Finally, while our results support a strong correlation between improved language neutrality and enhanced crosslingual transfer, we acknowledge that this relationship is not necessarily causal. 
However, prior studies have shown that improved language neutrality or alignment does not always yield better downstream outcomes \citep{gaschi-etal-2023-exploring,hua-etal-2024-mothello,liu-etal-2025-transliterations}, suggesting that language neutrality alone may not be a sufficient condition. 
We believe further research is needed to understand when and how language-neutral representations contribute effectively to transfer, which remains an open and important question for the multilingual NLP community.

\section*{Acknowledgments}
This work was funded by Deutsche Forschungsgemeinschaft
(project SCHU 2246/14-1)
and The European Research Council (NonSequeToR, grant \#740516).

\bibliography{anthology,custom}

\appendix

\section{Settings and Hyperparameters}
\seclabel{hyperparam}

We show the information of the evaluation datasets and used measures in Table \ref{tab:evaluation_info_task} and introduce the detailed settings and hyperparameters as follows.

\paragraph{Sentence Retrieval} We use English-aligned sentences (up to 500 and 1000 for SR-B and SR-T, respectively) from languages covered by Glot500-c \citep{imanigooghari-etal-2023-glot500}. No fine-tuning is needed for this evaluation type: we directly use each model as a text encoder and generate the sentence-level representation by averaging the contextual token embeddings at the \textbf{8th} layer, similar to previous work \citep{jalili-sabet-etal-2020-simalign,imanigooghari-etal-2023-glot500,liu-etal-2024-ofa}. We perform retrieval by sorting the pairwise similarities.

\paragraph{Text Classification} We add a 6-class or 7-class (for Taxi1500 and SIB200, respectively) sequence-level classification head onto the backbone model (no language or script IDs are required as input since the language modeling head is not needed in this sequence-level classification model). By default, we train the model on the English train set and store the best checkpoint on the English validation set. We train all models using AdamW optimizer \citep{ba2015adam, loshchilov2018decoupled} for a maximum of 40 epochs, with a learning rate of 1e-5 and an effective batch size of 16 (batch size of 8, gradient accumulation of 2). We use a single GTX 1080 Ti GPU for training. The evaluation is done in zero-shot transfer: we directly apply the best checkpoint to the test sets of all other languages.

\begin{table}
    \setlength{\belowcaptionskip}{-0.4cm}
  \small
	\centering
	\def\tablesep{0.05cm}
\begin{tabular}{
  @{\hspace{\tablesep}}l@{\hspace{\tablesep}}|
  @{\hspace{\tablesep}}r@{\hspace{\tablesep}}
  @{\hspace{\tablesep}}r@{\hspace{\tablesep}}
  @{\hspace{\tablesep}}r@{\hspace{\tablesep}}
  @{\hspace{\tablesep}}c@{\hspace{\tablesep}}
  @{\hspace{\tablesep}}r@{\hspace{\tablesep}}
  @{\hspace{\tablesep}}c@{\hspace{\tablesep}}
}
\toprule
 & |head| & |tail| & |Latn| & |non-Latn| & \#class & measure (\%) \\
  \midrule
SR-B    & 94 & 275 & 290 & 79 & - & top-10 Acc. \\
SR-T    & 70 & 28 & 64 & 34 & - & top-10 Acc. \\
Taxi1500    & 89 & 262 & 281 & 70 & 6 & F1 score \\
SIB200 & 78 & 94 & 117 & 55 & 7 & F1 score \\
NER & 89 & 75 & 104 & 60 & 7 & F1 score \\
POS & 63 & 28 & 57 & 34 & 18 & F1 score \\
\bottomrule
  \end{tabular}
  \caption{Information of the evaluation datasets and used measures. |head| (resp. |tail|): number of head (resp. tail) language-scripts. |Latn| (resp. |non-Latn|): number of languages written in Latin script (resp. non-Latn scripts). \#class: the number of the categories if it belongs to a text classification or sequence labeling task.}
  \label{tab:evaluation_info_task}
\end{table}

\paragraph{Sequence Labeling} We add a 7-class or 18-class (for NER and POS, respectively) token-level classification head onto the backbone model (no language or script IDs are required as input since the language modeling head is not needed in this token-level classification model). Similarly, we train the model on the English train set and store the best checkpoint on the English validation set by default. We train all models using the AdamW optimizer \citep{ba2015adam, loshchilov2018decoupled} for a maximum of 10 epochs. The learning rate is set to 2e-5, and the effective batch size is set to 32 (batch size of 8, gradient accumulation of 4). The training is done on a single GTX 1080 Ti GPU. The evaluation is done in zero-shot transfer: we directly apply the best checkpoint to the test sets of all other languages.

\begin{figure*}
    \centering
        \setlength{\belowcaptionskip}{-0.4cm}
    \includegraphics[width=0.95\textwidth]{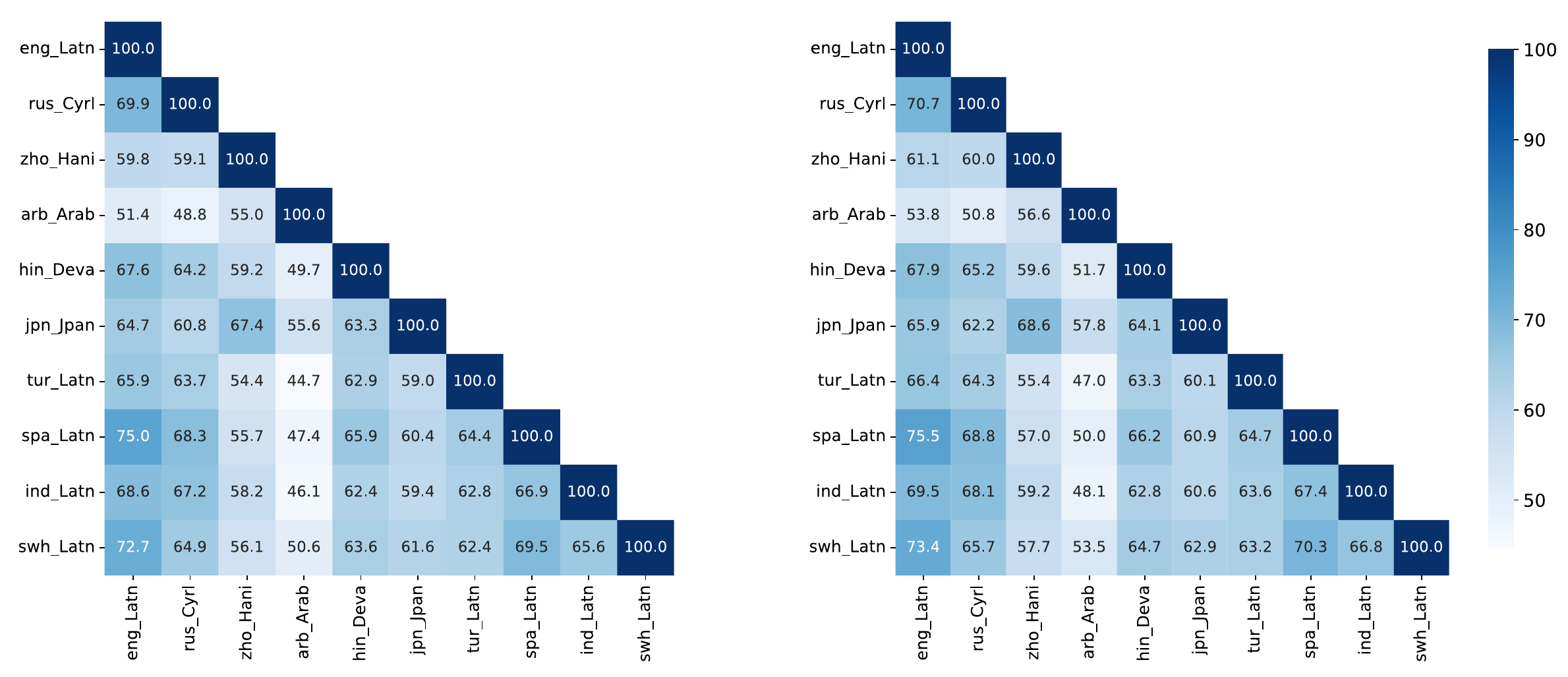}
    \caption{Comparison between baseline (left) and \frameworkname (right) in terms of the pairwise cosine similarity. \frameworkname achieves better similarity for each pair, indicating improved language neutrality of the representations.}
    \label{fig:sim_comparison}
\end{figure*}

\section{Pairwise Cosine Similarity}\seclabel{appendix:sim}

As introduced in \secref{Similarity}, we select 10 topologically different languages that are written in diverse scripts to assess the language neutrality: \textbf{eng\_Latn},
\textbf{rus\_Cyrl},
\textbf{zho\_Hani},
\textbf{arb\_Arab},
\textbf{hin\_Deva},
\textbf{jpn\_Jpan},
\textbf{tur\_Latn},
\textbf{spa\_Latn},
\textbf{ind\_Latn},
and
\textbf{swa\_Latn}.
We report the pairwise cosine similarity for the baseline and \frameworkname in Figure \ref{fig:sim_comparison}.

It can be observed that the similarity between any two languages in \frameworkname is consistently higher than in the baseline.
The absolute increase is small in general, due to the fact that (1) without the introduction of the auxiliary language and script embeddings, the baseline already assigns good similarity to translations and (2) \frameworkname does not introduce any additional parallel data in the pretraining, which is usually regarded as important to improve the similarity. 
Nevertheless, the consistent improvement indicates that \frameworkname effectively improves the language neutrality by decoupling language- and script-specific features into auxiliary embeddings.

\section{Results for Each Language Family}
We report the aggregated results for each language family for each task in Table \ref{tab:family}. We see consistent improvement for all language families in sentence retrieval and text classification tasks. For sequence tagging tasks, \frameworkname achieves similar performance compared with the baseline. This trend is similar to the main results we report in \secref{results}.

\begin{table*}[ht]
    \scriptsize
    \centering
    \setlength{\belowcaptionskip}{-0.2cm}
    \setlength{\tabcolsep}{1.0mm}{}
    \begin{tabular}{lrrrrrrrrr}
\toprule
& \multicolumn{8}{c}{\textbf{SR-B}}\\
\toprule
 & (indo1319, 93) & (atla1278, 69) & (aust1307, 55) & (turk1311, 23) & (sino1245, 23) & (maya1287, 15) & (afro1255, 12) & (other, 79) & (all, 369) \\
\midrule
Baseline & 61.4 & 37.3 & 42.9 & 60.9 & 31.6 & 15.5 & 29.5 & 31.3 & 42.9 \\
\frameworkname & \textbf{62.0} & \textbf{40.2} & \textbf{45.1} & \textbf{63.3} & \textbf{34.8} & \textbf{15.7} & \textbf{32.0} & \textbf{34.6} & \textbf{45.1} \\
\midrule
\midrule
& \multicolumn{8}{c}{\textbf{SR-T}}\\
\midrule
 & (indo1319, 54) & (atla1278, 2) & (aust1307, 7) & (turk1311, 7) & (sino1245, 3) & (maya1287, 0) & (afro1255, 5) & (other, 20) & (all, 98) \\
\midrule
Baseline & 74.2 & 50.0 & 48.7 & 71.3 & 81.7 & - & 52.1 & 68.7 & 69.7 \\
\frameworkname & \textbf{75.2} & \textbf{50.6} & \textbf{50.2} & \textbf{74.6} & \textbf{83.0} & - & \textbf{54.2} & \textbf{70.5} & \textbf{71.1} \\
\midrule
\midrule
& \multicolumn{8}{c}{\textbf{Taxi1500}}\\
\midrule
 & (indo1319, 87) & (atla1278, 68) & (aust1307, 51) & (turk1311, 18) & (sino1245, 22) & (maya1287, 15) & (afro1255, 11) & (other, 79) & (all, 351) \\
\midrule
Baseline & 60.8 & 42.6 & 51.6 & 60.3 & 49.5 & 42.4 & 35.7 & 46.1 & 50.3 \\
\frameworkname & \textbf{62.6} & \textbf{46.8} & \textbf{55.2} & \textbf{62.7} & \textbf{53.6} & \textbf{45.5} & \textbf{38.8} & \textbf{49.1} & \textbf{53.4} \\
\midrule
\midrule
& \multicolumn{8}{c}{\textbf{SIB200}}\\
\midrule
 & (indo1319, 71) & (atla1278, 33) & (aust1307, 17) & (turk1311, 10) & (sino1245, 5) & (maya1287, 0) & (afro1255, 13) & (other, 23) & (all, 172) \\
\midrule
Baseline & 82.1 & 59.0 & 76.4 & 80.5 & 67.4 & - & 73.0 & 75.1 & 75.0 \\
\frameworkname & \textbf{82.7} & \textbf{60.5} & \textbf{78.0} & \textbf{81.8} & \textbf{68.7} & - & \textbf{73.1} & \textbf{75.7} & \textbf{75.9} \\
\midrule
\midrule
& \multicolumn{8}{c}{\textbf{NER}}\\
\midrule
 & (indo1319, 94) & (atla1278, 5) & (aust1307, 12) & (turk1311, 12) & (sino1245, 7) & (maya1287, 0) & (afro1255, 6) & (other, 28) & (all, 164) \\
\midrule
Baseline & 66.8 & 60.4 & 58.8 & \textbf{62.1} & \textbf{37.1} & - & 53.8 & 56.7 & 62.2 \\
\frameworkname & \textbf{67.2} & \textbf{61.2} & \textbf{59.5} & 61.3 & 36.6 & - & \textbf{55.5} & \textbf{57.2} & \textbf{62.6} \\
\midrule
\midrule
& \multicolumn{8}{c}{\textbf{POS}}\\
\midrule
 & (indo1319, 54) & (atla1278, 2) & (aust1307, 4) & (turk1311, 5) & (sino1245, 3) & (maya1287, 1) & (afro1255, 6) & (other, 16) & (all, 91) \\
\midrule
Baseline & \textbf{78.0} & \textbf{61.9} & \textbf{74.4} & \textbf{72.1} & 33.3 & \textbf{61.1} & 64.1 & 60.3 & 71.5 \\
\frameworkname & \textbf{78.0} & 61.0 & \textbf{74.4} & 71.9 & \textbf{35.8} & 58.8 & \textbf{64.9} & \textbf{60.6} & \textbf{71.6} \\
\bottomrule
    \end{tabular}
    \caption{Aggregated performance of the baseline and \frameworkname for 7 major language families on all tasks. We report the average performance for \textbf{indo1319} (Indo-European), \textbf{atla1278 } (Atlantic-Congo), \textbf{aust1307} (Austronesian), \textbf{turk1311} (Turkic), \textbf{sino1245} (Sino-Tibetan), \textbf{maya1287} (Mayan), and \textbf{afro1255} (Afro-Asiatic). We classify the remaining languages into the group ``\textbf{other}''. In addition, we report the average over all languages (group ``\textbf{all}'').  The number of languages in that family is shown in parentheses.
    \textbf{Bold}: best result for each task.}
    \label{tab:family}
\end{table*}

\section{Complete Crosslingual Transfer Results}
\seclabel{complete}
We report the complete results of English-centric zero-shot crosslingual performance of baseline and \frameworkname for all tasks and languages in Table \ref{tab:srb_table1}, \ref{tab:srb_table2} (\textbf{SR-B}), Table \ref{tab:srt_table1} (\textbf{SR-T}), Table \ref{tab:taxi_table1}, \ref{tab:taxi_table2}(\textbf{Taxi1500}), \ref{tab:sib200_table1} (\textbf{SIB200}), Table \ref{tab:ner_table1} (\textbf{NER}), and Table \ref{tab:pos_table1} (\textbf{POS}). Each result is the average over fine-tuning the baseline or \frameworkname under five random seeds.

\input{bible}
\input{tatoeba}
\input{taxi1500}
\input{sib200}
\input{ner}
\input{pos}

\section{Transfer Results Using English and Closest Donor Language}\seclabel{donor_complete}
We report the complete results of the zero-shot crosslingual performance of \frameworkname when using English and the closest donor language as the source language in Table \ref{tab:taxi_donor_table1}, \ref{tab:taxi_donor_table2} (\textbf{Taxi1500}), \ref{tab:sib200_donor_table1} (\textbf{SIB200}), Table \ref{tab:ner_donor_table1} (\textbf{NER}), and Table \ref{tab:pos_donor_table1} (\textbf{POS}). Each result is directly obtained from a single run. We fine-tune the \frameworkname using different donor languages under the same random seed.

\input{taxi1500_donor}
\input{sib200_donor}
\input{ner_donor}
\input{pos_donor}

\end{document}

%% file: bible.tex
\begin{table*}
\centering
\setlength{\tabcolsep}{0.7mm}{}
\resizebox{\textwidth}{!}{
    \begin{tabular}{lrr|lrr|lrr|lrr}
    \toprule
    Language & Baseline & \frameworkname &     Language & Baseline & \frameworkname &     Language & Baseline & \frameworkname &     Language & Baseline & \frameworkname \\
    \midrule
ace\_Latn &  43.8 & \textbf{49.4} &ach\_Latn &  37.6 & \textbf{40.6} &acr\_Latn &  17.6 & \textbf{18.6} &afr\_Latn &  \textbf{74.2} & 72.4\\
agw\_Latn &  31.0 & \textbf{38.2} &ahk\_Latn &  3.4 & \textbf{3.8} &aka\_Latn &  41.8 & \textbf{48.4} &aln\_Latn &  \textbf{70.0} & \textbf{70.0}\\
als\_Latn &  \textbf{54.4} & \textbf{54.4} &alt\_Cyrl &  53.8 & \textbf{57.0} &alz\_Latn &  36.2 & \textbf{37.4} &amh\_Ethi &  44.4 & \textbf{51.2}\\
aoj\_Latn &  15.6 & \textbf{18.6} &arb\_Arab &  9.6 & \textbf{11.6} &arn\_Latn &  18.2 & \textbf{23.0} &ary\_Arab &  11.2 & \textbf{13.0}\\
arz\_Arab &  \textbf{15.2} & \textbf{15.2} &asm\_Beng &  \textbf{59.2} & 59.0 &ayr\_Latn &  37.6 & \textbf{46.0} &azb\_Arab &  55.6 & \textbf{59.0}\\
aze\_Latn &  73.4 & \textbf{75.4} &bak\_Cyrl &  58.8 & \textbf{62.2} &bam\_Latn &  38.4 & \textbf{44.8} &ban\_Latn &  33.0 & \textbf{33.2}\\
bar\_Latn &  32.2 & \textbf{34.0} &bba\_Latn &  26.2 & \textbf{31.0} &bbc\_Latn &  \textbf{60.8} & 58.8 &bci\_Latn &  \textbf{12.0} & 11.8\\
bcl\_Latn &  75.4 & \textbf{79.0} &bel\_Cyrl &  \textbf{70.6} & 69.6 &bem\_Latn &  51.0 & \textbf{54.4} &ben\_Beng &  53.4 & \textbf{55.4}\\
bhw\_Latn &  28.4 & \textbf{30.6} &bim\_Latn &  31.4 & \textbf{42.8} &bis\_Latn &  45.2 & \textbf{50.8} &bod\_Tibt &  29.6 & \textbf{33.6}\\
bqc\_Latn &  27.4 & \textbf{29.2} &bre\_Latn &  \textbf{31.8} & 30.0 &bts\_Latn &  \textbf{62.4} & 62.0 &btx\_Latn &  \textbf{57.2} & 55.8\\
bul\_Cyrl &  79.8 & \textbf{80.0} &bum\_Latn &  32.8 & \textbf{35.2} &bzj\_Latn &  69.8 & \textbf{70.2} &cab\_Latn &  11.6 & \textbf{11.8}\\
cac\_Latn &  10.8 & \textbf{11.8} &cak\_Latn &  \textbf{17.8} & 16.6 &caq\_Latn &  26.0 & \textbf{29.8} &cat\_Latn &  \textbf{85.4} & 83.2\\
cbk\_Latn &  54.8 & \textbf{56.2} &cce\_Latn &  41.8 & \textbf{45.4} &ceb\_Latn &  70.4 & \textbf{70.6} &ces\_Latn &  \textbf{68.2} & 67.0\\
cfm\_Latn &  34.4 & \textbf{38.8} &che\_Cyrl &  10.2 & \textbf{11.2} &chk\_Latn &  35.2 & \textbf{43.0} &chv\_Cyrl &  45.0 & \textbf{54.4}\\
ckb\_Arab &  31.2 & \textbf{32.8} &cmn\_Hani &  \textbf{41.4} & 40.8 &cnh\_Latn &  38.2 & \textbf{43.2} &crh\_Cyrl &  67.2 & \textbf{70.0}\\
crs\_Latn &  \textbf{85.6} & 84.4 &csy\_Latn &  40.2 & \textbf{49.6} &ctd\_Latn &  44.4 & \textbf{50.6} &ctu\_Latn &  \textbf{16.6} & 16.0\\
cuk\_Latn &  \textbf{17.0} & \textbf{17.0} &cym\_Latn &  \textbf{45.6} & 43.8 &dan\_Latn &  \textbf{72.4} & 71.8 &deu\_Latn &  73.8 & \textbf{74.0}\\
djk\_Latn &  \textbf{38.0} & \textbf{38.0} &dln\_Latn &  46.6 & \textbf{51.4} &dtp\_Latn &  17.0 & \textbf{17.8} &dyu\_Latn &  33.0 & \textbf{40.2}\\
dzo\_Tibt &  28.4 & \textbf{33.0} &efi\_Latn &  41.6 & \textbf{53.6} &ell\_Grek &  48.2 & \textbf{49.2} &enm\_Latn &  \textbf{69.4} & \textbf{69.4}\\
epo\_Latn &  \textbf{67.4} & 65.8 &est\_Latn &  \textbf{66.4} & 66.0 &eus\_Latn &  23.8 & \textbf{24.2} &ewe\_Latn &  33.2 & \textbf{34.8}\\
fao\_Latn &  \textbf{79.8} & 78.4 &fas\_Arab &  80.2 & \textbf{84.2} &fij\_Latn &  30.0 & \textbf{31.0} &fil\_Latn &  \textbf{77.6} & 77.2\\
fin\_Latn &  65.4 & \textbf{66.0} &fon\_Latn &  20.2 & \textbf{25.2} &fra\_Latn &  \textbf{87.4} & 87.2 &fry\_Latn &  \textbf{47.0} & 44.0\\
gaa\_Latn &  34.4 & \textbf{40.6} &gil\_Latn &  30.0 & \textbf{31.6} &giz\_Latn &  32.4 & \textbf{36.4} &gkn\_Latn &  20.4 & \textbf{24.2}\\
gkp\_Latn &  13.2 & \textbf{14.6} &gla\_Latn &  \textbf{39.0} & 38.0 &gle\_Latn &  \textbf{41.2} & 38.4 &glv\_Latn &  37.2 & \textbf{38.6}\\
gom\_Latn &  33.2 & \textbf{36.0} &gor\_Latn &  21.8 & \textbf{23.0} &grc\_Grek &  44.4 & \textbf{47.0} &guc\_Latn &  \textbf{9.8} & 8.2\\
gug\_Latn &  28.2 & \textbf{31.2} &guj\_Gujr &  \textbf{69.8} & 67.6 &gur\_Latn &  17.6 & \textbf{18.2} &guw\_Latn &  36.8 & \textbf{45.4}\\
gya\_Latn &  27.6 & \textbf{32.6} &gym\_Latn &  \textbf{13.6} & 13.0 &hat\_Latn &  \textbf{76.4} & 74.6 &hau\_Latn &  57.6 & \textbf{59.6}\\
haw\_Latn &  28.0 & \textbf{30.4} &heb\_Hebr &  21.6 & \textbf{23.0} &hif\_Latn &  33.2 & \textbf{34.6} &hil\_Latn &  74.0 & \textbf{79.8}\\
hin\_Deva &  \textbf{75.6} & 74.6 &hin\_Latn &  34.2 & \textbf{36.2} &hmo\_Latn &  44.2 & \textbf{57.0} &hne\_Deva &  71.6 & \textbf{73.6}\\
hnj\_Latn &  39.6 & \textbf{46.6} &hra\_Latn &  43.4 & \textbf{46.4} &hrv\_Latn &  \textbf{80.4} & 79.8 &hui\_Latn &  19.8 & \textbf{22.0}\\
hun\_Latn &  65.6 & \textbf{69.0} &hus\_Latn &  14.8 & \textbf{16.2} &hye\_Armn &  62.8 & \textbf{65.6} &iba\_Latn &  70.2 & \textbf{71.6}\\
ibo\_Latn &  \textbf{32.4} & 31.6 &ifa\_Latn &  26.2 & \textbf{29.0} &ifb\_Latn &  \textbf{28.6} & \textbf{28.6} &ikk\_Latn &  30.2 & \textbf{46.4}\\
ilo\_Latn &  53.4 & \textbf{54.4} &ind\_Latn &  78.4 & \textbf{78.6} &isl\_Latn &  71.0 & \textbf{71.8} &ita\_Latn &  76.2 & \textbf{76.8}\\
ium\_Latn &  20.0 & \textbf{23.2} &ixl\_Latn &  13.8 & \textbf{14.4} &izz\_Latn &  19.6 & \textbf{22.6} &jam\_Latn &  \textbf{61.0} & 59.2\\
jav\_Latn &  \textbf{55.4} & 52.0 &jpn\_Jpan &  65.8 & \textbf{67.6} &kaa\_Cyrl &  71.2 & \textbf{75.0} &kaa\_Latn &  32.0 & \textbf{37.6}\\
kab\_Latn &  12.2 & \textbf{13.4} &kac\_Latn &  22.2 & \textbf{27.0} &kal\_Latn &  12.6 & \textbf{16.8} &kan\_Knda &  50.0 & \textbf{52.8}\\
kat\_Geor &  49.6 & \textbf{52.4} &kaz\_Cyrl &  69.4 & \textbf{70.4} &kbp\_Latn &  21.8 & \textbf{26.8} &kek\_Latn &  16.6 & \textbf{18.6}\\
khm\_Khmr &  39.4 & \textbf{43.0} &kia\_Latn &  24.6 & \textbf{28.8} &kik\_Latn &  44.4 & \textbf{48.4} &kin\_Latn &  56.6 & \textbf{60.2}\\
kir\_Cyrl &  69.8 & \textbf{70.2} &kjb\_Latn &  23.4 & \textbf{26.0} &kjh\_Cyrl &  45.6 & \textbf{50.6} &kmm\_Latn &  33.8 & \textbf{38.0}\\
kmr\_Cyrl &  \textbf{42.0} & 40.2 &kmr\_Latn &  60.2 & \textbf{60.4} &knv\_Latn &  7.0 & \textbf{8.4} &kor\_Hang &  60.8 & \textbf{64.0}\\
kpg\_Latn &  42.6 & \textbf{48.8} &krc\_Cyrl &  59.8 & \textbf{62.2} &kri\_Latn &  61.4 & \textbf{62.6} &ksd\_Latn &  31.4 & \textbf{41.0}\\
kss\_Latn &  5.2 & \textbf{6.0} &ksw\_Mymr &  26.2 & \textbf{28.0} &kua\_Latn &  43.0 & \textbf{43.8} &lam\_Latn &  20.4 & \textbf{22.8}\\
lao\_Laoo &  41.6 & \textbf{47.2} &lat\_Latn &  56.6 & \textbf{58.0} &lav\_Latn &  69.8 & \textbf{71.2} &ldi\_Latn &  \textbf{22.4} & 22.0\\
leh\_Latn &  \textbf{46.8} & 45.8 &lhu\_Latn &  \textbf{4.4} & 4.2 &lin\_Latn &  64.6 & \textbf{71.0} &lit\_Latn &  \textbf{67.0} & 66.6\\
loz\_Latn &  \textbf{46.8} & 45.6 &ltz\_Latn &  \textbf{63.8} & 63.2 &lug\_Latn &  37.2 & \textbf{40.8} &luo\_Latn &  \textbf{42.8} & 42.6\\
lus\_Latn &  46.6 & \textbf{53.2} &lzh\_Hani &  59.8 & \textbf{62.4} &mad\_Latn &  42.6 & \textbf{44.6} &mah\_Latn &  30.4 & \textbf{33.8}\\
mai\_Deva &  52.6 & \textbf{56.0} &mal\_Mlym &  51.6 & \textbf{57.4} &mam\_Latn &  \textbf{10.2} & \textbf{10.2} &mar\_Deva &  68.4 & \textbf{71.4}\\
mau\_Latn &  2.8 & \textbf{3.4} &mbb\_Latn &  22.0 & \textbf{29.8} &mck\_Latn &  \textbf{55.6} & 53.4 &mcn\_Latn &  34.2 & \textbf{40.8}\\
mco\_Latn &  \textbf{6.6} & 6.4 &mdy\_Ethi &  21.4 & \textbf{30.6} &meu\_Latn &  48.8 & \textbf{52.0} &mfe\_Latn &  \textbf{77.4} & \textbf{77.4}\\
    \bottomrule
    \end{tabular}
}
    \caption{Top-10 accuracy of models on \textbf{SR-B} (Part I).}\label{tab:srb_table1}
\end{table*}

\begin{table*}
\centering
\setlength{\tabcolsep}{0.7mm}{}
\resizebox{\textwidth}{!}{
    \begin{tabular}{lrr|lrr|lrr|lrr}
    \toprule
    Language & Baseline & \frameworkname &     Language & Baseline & \frameworkname &     Language & Baseline & \frameworkname &     Language & Baseline & \frameworkname \\
    \midrule
    mgh\_Latn &  17.4 & \textbf{20.8} &mgr\_Latn &  \textbf{48.6} & 47.2 &mhr\_Cyrl &  37.4 & \textbf{43.2} &min\_Latn &  \textbf{32.4} & 29.6\\
miq\_Latn &  28.8 & \textbf{36.8} &mkd\_Cyrl &  78.4 & \textbf{78.8} &mlg\_Latn &  60.2 & \textbf{61.2} &mlt\_Latn &  48.0 & \textbf{50.4}\\
mos\_Latn &  32.2 & \textbf{32.8} &mps\_Latn &  16.4 & \textbf{20.6} &mri\_Latn &  45.6 & \textbf{55.0} &mrw\_Latn &  34.0 & \textbf{40.6}\\
msa\_Latn &  43.6 & \textbf{44.2} &mwm\_Latn &  24.0 & \textbf{25.6} &mxv\_Latn &  \textbf{7.0} & \textbf{7.0} &mya\_Mymr &  25.8 & \textbf{28.0}\\
myv\_Cyrl &  26.6 & \textbf{30.6} &mzh\_Latn &  24.6 & \textbf{25.4} &nan\_Latn &  13.2 & \textbf{13.6} &naq\_Latn &  16.8 & \textbf{26.8}\\
nav\_Latn &  \textbf{8.6} & \textbf{8.6} &nbl\_Latn &  \textbf{49.4} & 48.4 &nch\_Latn &  \textbf{21.6} & \textbf{21.6} &ncj\_Latn &  18.8 & \textbf{19.4}\\
ndc\_Latn &  32.4 & \textbf{36.2} &nde\_Latn &  51.0 & \textbf{54.8} &ndo\_Latn &  41.0 & \textbf{44.0} &nds\_Latn &  \textbf{38.4} & \textbf{38.4}\\
nep\_Deva &  56.4 & \textbf{59.0} &ngu\_Latn &  \textbf{26.2} & 26.0 &nia\_Latn &  25.6 & \textbf{28.0} &nld\_Latn &  \textbf{78.4} & 78.0\\
nmf\_Latn &  25.6 & \textbf{28.2} &nnb\_Latn &  33.2 & \textbf{38.8} &nno\_Latn &  \textbf{76.8} & 75.8 &nob\_Latn &  \textbf{85.4} & 85.0\\
nor\_Latn &  \textbf{85.8} & 83.4 &npi\_Deva &  77.4 & \textbf{80.8} &nse\_Latn &  48.4 & \textbf{51.8} &nso\_Latn &  46.2 & \textbf{50.2}\\
nya\_Latn &  \textbf{57.6} & \textbf{57.6} &nyn\_Latn &  \textbf{48.8} & 47.4 &nyy\_Latn &  23.4 & \textbf{24.6} &nzi\_Latn &  29.2 & \textbf{34.4}\\
ori\_Orya &  51.2 & \textbf{53.4} &ory\_Orya &  46.4 & \textbf{49.8} &oss\_Cyrl &  41.4 & \textbf{56.4} &ote\_Latn &  12.0 & \textbf{13.2}\\
pag\_Latn &  \textbf{55.2} & 52.2 &pam\_Latn &  37.4 & \textbf{41.2} &pan\_Guru &  \textbf{46.2} & 45.4 &pap\_Latn &  72.8 & \textbf{75.0}\\
pau\_Latn &  17.0 & \textbf{23.4} &pcm\_Latn &  \textbf{69.8} & 69.4 &pdt\_Latn &  \textbf{69.4} & 66.0 &pes\_Arab &  74.2 & \textbf{75.2}\\
pis\_Latn &  51.4 & \textbf{54.8} &pls\_Latn &  27.0 & \textbf{31.8} &plt\_Latn &  60.2 & \textbf{60.8} &poh\_Latn &  10.6 & \textbf{11.4}\\
pol\_Latn &  73.8 & \textbf{75.6} &pon\_Latn &  21.4 & \textbf{24.0} &por\_Latn &  \textbf{81.8} & 81.0 &prk\_Latn &  42.0 & \textbf{47.4}\\
prs\_Arab &  84.6 & \textbf{87.0} &pxm\_Latn &  18.2 & \textbf{19.8} &qub\_Latn &  30.6 & \textbf{35.6} &quc\_Latn &  \textbf{18.6} & 17.4\\
qug\_Latn &  53.6 & \textbf{59.2} &quh\_Latn &  40.2 & \textbf{43.8} &quw\_Latn &  46.2 & \textbf{50.4} &quy\_Latn &  47.4 & \textbf{54.4}\\
quz\_Latn &  59.4 & \textbf{63.6} &qvi\_Latn &  49.2 & \textbf{57.6} &rap\_Latn &  17.0 & \textbf{17.8} &rar\_Latn &  \textbf{20.4} & 19.8\\
rmy\_Latn &  30.4 & \textbf{32.2} &ron\_Latn &  \textbf{69.4} & 69.0 &rop\_Latn &  35.8 & \textbf{41.4} &rug\_Latn &  37.8 & \textbf{38.4}\\
run\_Latn &  48.2 & \textbf{52.4} &rus\_Cyrl &  74.6 & \textbf{76.4} &sag\_Latn &  39.6 & \textbf{45.4} &sah\_Cyrl &  43.4 & \textbf{45.8}\\
san\_Deva &  \textbf{24.2} & 23.6 &san\_Latn &  \textbf{7.8} & 7.4 &sba\_Latn &  28.0 & \textbf{29.2} &seh\_Latn &  67.4 & \textbf{69.4}\\
sin\_Sinh &  45.6 & \textbf{49.0} &slk\_Latn &  \textbf{69.8} & 69.2 &slv\_Latn &  \textbf{61.2} & 60.8 &sme\_Latn &  35.0 & \textbf{37.6}\\
smo\_Latn &  27.6 & \textbf{28.8} &sna\_Latn &  38.4 & \textbf{41.2} &snd\_Arab &  \textbf{67.2} & 65.0 &som\_Latn &  \textbf{35.0} & 34.8\\
sop\_Latn &  \textbf{32.4} & 28.8 &sot\_Latn &  48.4 & \textbf{52.4} &spa\_Latn &  80.8 & \textbf{81.4} &sqi\_Latn &  62.2 & \textbf{64.8}\\
srm\_Latn &  \textbf{28.2} & 26.6 &srn\_Latn &  75.4 & \textbf{75.6} &srp\_Cyrl &  \textbf{87.2} & 85.8 &srp\_Latn &  \textbf{85.8} & 85.4\\
ssw\_Latn &  42.8 & \textbf{47.0} &sun\_Latn &  52.0 & \textbf{54.0} &suz\_Deva &  21.0 & \textbf{22.6} &swe\_Latn &  \textbf{78.6} & 77.0\\
swh\_Latn &  \textbf{71.6} & 71.4 &sxn\_Latn &  20.6 & \textbf{20.8} &tam\_Taml &  47.0 & \textbf{50.6} &tat\_Cyrl &  68.2 & \textbf{70.4}\\
tbz\_Latn &  13.2 & \textbf{18.2} &tca\_Latn &  10.0 & \textbf{13.8} &tdt\_Latn &  50.0 & \textbf{53.6} &tel\_Telu &  48.0 & \textbf{50.2}\\
teo\_Latn &  19.4 & \textbf{19.6} &tgk\_Cyrl &  69.2 & \textbf{69.4} &tgl\_Latn &  \textbf{79.6} & 78.0 &tha\_Thai &  33.8 & \textbf{38.0}\\
tih\_Latn &  42.2 & \textbf{46.4} &tir\_Ethi &  32.2 & \textbf{34.8} &tlh\_Latn &  62.0 & \textbf{66.4} &tob\_Latn &  \textbf{11.6} & 11.4\\
toh\_Latn &  36.8 & \textbf{41.8} &toi\_Latn &  \textbf{39.4} & \textbf{39.4} &toj\_Latn &  \textbf{14.8} & 12.6 &ton\_Latn &  16.0 & \textbf{16.6}\\
top\_Latn &  \textbf{6.6} & 6.0 &tpi\_Latn &  58.0 & \textbf{62.2} &tpm\_Latn &  \textbf{27.4} & 23.0 &tsn\_Latn &  32.6 & \textbf{34.6}\\
tso\_Latn &  50.0 & \textbf{51.0} &tsz\_Latn &  21.2 & \textbf{25.8} &tuc\_Latn &  25.6 & \textbf{32.4} &tui\_Latn &  29.8 & \textbf{31.0}\\
tuk\_Cyrl &  67.4 & \textbf{69.4} &tuk\_Latn &  67.6 & \textbf{70.0} &tum\_Latn &  \textbf{58.4} & 57.0 &tur\_Latn &  70.2 & \textbf{70.4}\\
twi\_Latn &  35.0 & \textbf{42.0} &tyv\_Cyrl &  \textbf{44.2} & 43.4 &tzh\_Latn &  19.0 & \textbf{19.8} &tzo\_Latn &  \textbf{14.2} & 13.6\\
udm\_Cyrl &  41.6 & \textbf{45.2} &uig\_Arab &  47.4 & \textbf{50.8} &uig\_Latn &  57.2 & \textbf{58.8} &ukr\_Cyrl &  67.0 & \textbf{68.0}\\
urd\_Arab &  60.4 & \textbf{61.4} &uzb\_Cyrl &  80.6 & \textbf{81.2} &uzb\_Latn &  \textbf{70.0} & 68.2 &uzn\_Cyrl &  82.4 & \textbf{83.0}\\
ven\_Latn &  37.2 & \textbf{42.0} &vie\_Latn &  68.0 & \textbf{69.4} &wal\_Latn &  35.0 & \textbf{43.4} &war\_Latn &  42.6 & \textbf{44.0}\\
wbm\_Latn &  37.6 & \textbf{46.2} &wol\_Latn &  31.8 & \textbf{33.2} &xav\_Latn &  3.8 & \textbf{4.0} &xho\_Latn &  42.6 & \textbf{44.2}\\
yan\_Latn &  16.4 & \textbf{27.2} &yao\_Latn &  37.4 & \textbf{37.6} &yap\_Latn &  15.8 & \textbf{19.6} &yom\_Latn &  37.6 & \textbf{40.0}\\
yor\_Latn &  27.4 & \textbf{28.8} &yua\_Latn &  \textbf{13.2} & 12.8 &yue\_Hani &  \textbf{17.2} & \textbf{17.2} &zai\_Latn &  29.0 & \textbf{30.6}\\
zho\_Hani &  41.6 & \textbf{41.8} &zlm\_Latn &  \textbf{84.8} & \textbf{84.8} &zom\_Latn &  39.6 & \textbf{45.0} &zsm\_Latn &  90.0 & \textbf{91.0}\\
zul\_Latn &  \textbf{51.4} & 51.0 & & & & & &\\
    \bottomrule
    \end{tabular}
}
    \caption{Top-10 accuracy of models on \textbf{SR-B} (Part II).}\label{tab:srb_table2}
\end{table*}

%% file: tatoeba.tex
\begin{table*}
\centering
\setlength{\tabcolsep}{0.7mm}{}
\resizebox{\textwidth}{!}{
    \begin{tabular}{lrr|lrr|lrr|lrr}
    \toprule
    Language & Baseline & \frameworkname &     Language & Baseline & \frameworkname &     Language & Baseline & \frameworkname &     Language & Baseline & \frameworkname \\
    \midrule
    afr\_Latn &  77.9 & \textbf{80.4} &amh\_Ethi &  47.0 & \textbf{52.4} &ara\_Arab &  \textbf{69.4} & 68.7 &arz\_Arab &  61.8 & \textbf{63.9}\\
ast\_Latn &  80.3 & \textbf{84.3} &aze\_Latn &  82.6 & \textbf{84.1} &bel\_Cyrl &  \textbf{83.6} & 83.0 &ben\_Beng &  72.1 & \textbf{74.9}\\
bos\_Latn &  90.1 & \textbf{90.4} &bre\_Latn &  17.4 & \textbf{18.2} &bul\_Cyrl &  87.5 & \textbf{89.2} &cat\_Latn &  78.2 & \textbf{78.6}\\
cbk\_Latn &  \textbf{49.4} & 48.0 &ceb\_Latn &  39.0 & \textbf{42.5} &ces\_Latn &  \textbf{75.7} & 73.5 &cmn\_Hani &  87.1 & \textbf{87.4}\\
csb\_Latn &  38.3 & \textbf{38.7} &cym\_Latn &  52.2 & \textbf{55.0} &dan\_Latn &  91.7 & \textbf{92.9} &deu\_Latn &  95.5 & \textbf{95.7}\\
dtp\_Latn &  17.0 & \textbf{19.3} &ell\_Grek &  79.3 & \textbf{82.7} &epo\_Latn &  71.8 & \textbf{74.8} &est\_Latn &  68.2 & \textbf{69.9}\\
eus\_Latn &  52.2 & \textbf{55.4} &fao\_Latn &  \textbf{77.1} & 75.6 &fin\_Latn &  72.3 & \textbf{74.2} &fra\_Latn &  \textbf{85.3} & 85.2\\
fry\_Latn &  75.1 & \textbf{79.2} &gla\_Latn &  38.4 & \textbf{38.6} &gle\_Latn &  44.8 & \textbf{48.3} &glg\_Latn &  \textbf{77.1} & 76.4\\
gsw\_Latn &  58.1 & \textbf{63.2} &heb\_Hebr &  71.4 & \textbf{74.9} &hin\_Deva &  \textbf{88.1} & 87.3 &hrv\_Latn &  \textbf{87.9} & 87.5\\
hsb\_Latn &  \textbf{49.7} & \textbf{49.7} &hun\_Latn &  71.5 & \textbf{73.2} &hye\_Armn &  79.1 & \textbf{81.3} &ido\_Latn &  54.6 & \textbf{55.8}\\
ile\_Latn &  71.2 & \textbf{71.5} &ina\_Latn &  89.2 & \textbf{90.7} &ind\_Latn &  88.1 & \textbf{88.9} &isl\_Latn &  84.0 & \textbf{84.5}\\
ita\_Latn &  84.1 & \textbf{85.7} &jpn\_Jpan &  \textbf{77.2} & 77.1 &kab\_Latn &  10.8 & \textbf{11.0} &kat\_Geor &  71.2 & \textbf{72.4}\\
kaz\_Cyrl &  74.6 & \textbf{77.7} &khm\_Khmr &  57.5 & \textbf{63.0} &kor\_Hang &  80.8 & \textbf{81.1} &kur\_Latn &  49.8 & \textbf{52.4}\\
lat\_Latn &  39.2 & \textbf{42.1} &lfn\_Latn &  55.8 & \textbf{56.8} &lit\_Latn &  70.4 & \textbf{72.9} &lvs\_Latn &  76.2 & \textbf{78.1}\\
mal\_Mlym &  87.5 & \textbf{91.6} &mar\_Deva &  79.8 & \textbf{81.6} &mhr\_Cyrl &  27.7 & \textbf{33.4} &mkd\_Cyrl &  \textbf{79.6} & 79.4\\
mon\_Cyrl &  78.2 & \textbf{80.5} &nds\_Latn &  71.3 & \textbf{72.5} &nld\_Latn &  92.4 & \textbf{93.4} &nno\_Latn &  85.5 & \textbf{87.4}\\
nob\_Latn &  94.5 & \textbf{95.3} &oci\_Latn &  \textbf{46.6} & 44.9 &pam\_Latn &  \textbf{10.2} & \textbf{10.2} &pes\_Arab &  86.7 & \textbf{86.9}\\
pms\_Latn &  49.5 & \textbf{50.9} &pol\_Latn &  \textbf{84.3} & 83.4 &por\_Latn &  90.2 & \textbf{90.7} &ron\_Latn &  86.0 & \textbf{86.9}\\
rus\_Cyrl &  91.6 & \textbf{92.1} &slk\_Latn &  77.9 & \textbf{78.2} &slv\_Latn &  \textbf{76.2} & 75.9 &spa\_Latn &  \textbf{88.6} & 88.3\\
sqi\_Latn &  84.1 & \textbf{85.2} &srp\_Latn &  \textbf{89.7} & 89.6 &swe\_Latn &  89.4 & \textbf{89.6} &swh\_Latn &  \textbf{45.1} & 44.9\\
tam\_Taml &  \textbf{50.2} & 45.0 &tat\_Cyrl &  71.2 & \textbf{74.6} &tel\_Telu &  72.6 & \textbf{74.8} &tgl\_Latn &  73.9 & \textbf{74.2}\\
tha\_Thai &  75.4 & \textbf{79.2} &tuk\_Latn &  62.1 & \textbf{68.0} &tur\_Latn &  79.1 & \textbf{82.0} &uig\_Arab &  64.7 & \textbf{68.4}\\
ukr\_Cyrl &  84.9 & \textbf{86.5} &urd\_Arab &  78.5 & \textbf{81.7} &uzb\_Cyrl &  65.0 & \textbf{67.3} &vie\_Latn &  \textbf{88.9} & 88.8\\
war\_Latn &  22.7 & \textbf{25.2} &wuu\_Hani &  79.0 & \textbf{82.4} &xho\_Latn &  54.9 & \textbf{56.3} &yid\_Hebr &  65.8 & \textbf{67.6}\\
yue\_Hani &  79.0 & \textbf{79.3} &zsm\_Latn &  90.2 & \textbf{91.0} & & & &\\
    \bottomrule
    \end{tabular}
}
    \caption{Top-10 accuracy of models on \textbf{SR-T}.}\label{tab:srt_table1}
\end{table*}

%% file: taxi1500.tex
\begin{table*}
\centering
\setlength{\tabcolsep}{0.7mm}{}
\resizebox{\textwidth}{!}{
    \begin{tabular}{lrr|lrr|lrr|lrr}
    \toprule
    Language & Baseline & \frameworkname &     Language & Baseline & \frameworkname &     Language & Baseline & \frameworkname &     Language & Baseline & \frameworkname \\
    \midrule
ace\_Latn &  \textbf{66.6} & 64.5 &ach\_Latn &  36.6 & \textbf{40.5} &acr\_Latn &  45.6 & \textbf{51.4} &afr\_Latn &  \textbf{60.7} & 58.9\\
agw\_Latn &  53.6 & \textbf{55.8} &ahk\_Latn &  \textbf{7.6} & 7.3 &aka\_Latn &  43.3 & \textbf{47.4} &aln\_Latn &  56.0 & \textbf{57.4}\\
als\_Latn &  56.6 & \textbf{57.1} &alt\_Cyrl &  48.5 & \textbf{49.9} &alz\_Latn &  31.9 & \textbf{39.1} &amh\_Ethi &  \textbf{8.7} & 7.9\\
aoj\_Latn &  36.8 & \textbf{42.5} &arn\_Latn &  41.9 & \textbf{45.1} &ary\_Arab &  33.8 & \textbf{34.3} &arz\_Arab &  36.2 & \textbf{40.5}\\
asm\_Beng &  \textbf{63.2} & 62.4 &ayr\_Latn &  55.9 & \textbf{57.3} &azb\_Arab &  \textbf{63.7} & 62.8 &aze\_Latn &  67.3 & \textbf{70.3}\\
bak\_Cyrl &  \textbf{60.8} & 58.9 &bam\_Latn &  43.8 & \textbf{49.6} &ban\_Latn &  42.6 & \textbf{47.7} &bar\_Latn &  46.2 & \textbf{49.7}\\
bba\_Latn &  40.9 & \textbf{43.3} &bci\_Latn &  31.3 & \textbf{33.7} &bcl\_Latn &  55.0 & \textbf{61.7} &bel\_Cyrl &  \textbf{59.8} & \textbf{59.8}\\
bem\_Latn &  45.5 & \textbf{50.6} &ben\_Beng &  62.4 & \textbf{65.9} &bhw\_Latn &  45.3 & \textbf{53.2} &bim\_Latn &  49.4 & \textbf{49.8}\\
bis\_Latn &  67.2 & \textbf{71.8} &bqc\_Latn &  32.3 & \textbf{36.7} &bre\_Latn &  37.3 & \textbf{44.0} &btx\_Latn &  54.0 & \textbf{64.6}\\
bul\_Cyrl &  \textbf{65.3} & \textbf{65.3} &bum\_Latn &  39.5 & \textbf{45.3} &bzj\_Latn &  66.2 & \textbf{67.7} &cab\_Latn &  24.5 & \textbf{31.0}\\
cac\_Latn &  44.8 & \textbf{45.9} &cak\_Latn &  53.2 & \textbf{54.4} &caq\_Latn &  39.9 & \textbf{45.7} &cat\_Latn &  \textbf{63.6} & 61.6\\
cbk\_Latn &  62.2 & \textbf{68.4} &cce\_Latn &  42.5 & \textbf{48.3} &ceb\_Latn &  53.1 & \textbf{56.0} &ces\_Latn &  61.6 & \textbf{66.3}\\
cfm\_Latn &  55.3 & \textbf{64.8} &che\_Cyrl &  17.6 & \textbf{23.2} &chv\_Cyrl &  56.7 & \textbf{61.7} &cmn\_Hani &  67.8 & \textbf{69.2}\\
cnh\_Latn &  61.9 & \textbf{64.5} &crh\_Cyrl &  61.2 & \textbf{64.3} &crs\_Latn &  \textbf{65.9} & 64.3 &csy\_Latn &  53.6 & \textbf{62.8}\\
ctd\_Latn &  53.5 & \textbf{58.5} &ctu\_Latn &  \textbf{52.2} & 51.8 &cuk\_Latn &  40.2 & \textbf{43.3} &cym\_Latn &  \textbf{50.1} & 49.4\\
dan\_Latn &  62.4 & \textbf{63.6} &deu\_Latn &  53.4 & \textbf{56.7} &djk\_Latn &  46.5 & \textbf{54.7} &dln\_Latn &  49.3 & \textbf{61.4}\\
dtp\_Latn &  50.2 & \textbf{51.8} &dyu\_Latn &  48.0 & \textbf{57.1} &dzo\_Tibt &  55.9 & \textbf{57.8} &efi\_Latn &  52.4 & \textbf{56.2}\\
ell\_Grek &  59.8 & \textbf{61.3} &eng\_Latn &  74.3 & \textbf{75.3} &enm\_Latn &  \textbf{72.2} & 70.7 &epo\_Latn &  57.6 & \textbf{58.8}\\
est\_Latn &  56.8 & \textbf{57.5} &eus\_Latn &  23.3 & \textbf{27.8} &ewe\_Latn &  43.6 & \textbf{52.4} &fao\_Latn &  57.5 & \textbf{59.5}\\
fas\_Arab &  \textbf{71.7} & 70.3 &fij\_Latn &  44.2 & \textbf{48.5} &fil\_Latn &  57.5 & \textbf{58.8} &fin\_Latn &  58.3 & \textbf{59.2}\\
fon\_Latn &  42.9 & \textbf{44.0} &fra\_Latn &  65.4 & \textbf{70.4} &fry\_Latn &  40.0 & \textbf{43.1} &gaa\_Latn &  40.2 & \textbf{41.8}\\
gil\_Latn &  42.0 & \textbf{44.5} &giz\_Latn &  45.1 & \textbf{49.7} &gkn\_Latn &  38.3 & \textbf{43.7} &gkp\_Latn &  32.1 & \textbf{37.5}\\
gla\_Latn &  48.3 & \textbf{49.4} &gle\_Latn &  42.9 & \textbf{44.9} &glv\_Latn &  39.7 & \textbf{43.5} &gom\_Latn &  35.5 & \textbf{38.2}\\
gor\_Latn &  42.5 & \textbf{50.8} &guc\_Latn &  33.4 & \textbf{38.4} &gug\_Latn &  36.2 & \textbf{41.2} &guj\_Gujr &  68.8 & \textbf{70.3}\\
gur\_Latn &  34.1 & \textbf{43.3} &guw\_Latn &  48.4 & \textbf{52.1} &gya\_Latn &  \textbf{40.6} & 39.9 &gym\_Latn &  41.1 & \textbf{47.2}\\
hat\_Latn &  62.8 & \textbf{65.2} &hau\_Latn &  54.2 & \textbf{58.6} &haw\_Latn &  30.2 & \textbf{38.3} &heb\_Hebr &  18.7 & \textbf{21.7}\\
hif\_Latn &  46.2 & \textbf{47.9} &hil\_Latn &  65.1 & \textbf{67.3} &hin\_Deva &  66.2 & \textbf{69.0} &hmo\_Latn &  60.7 & \textbf{65.3}\\
hne\_Deva &  66.3 & \textbf{67.4} &hnj\_Latn &  63.4 & \textbf{66.3} &hra\_Latn &  51.5 & \textbf{56.0} &hrv\_Latn &  63.4 & \textbf{67.3}\\
hui\_Latn &  46.5 & \textbf{50.2} &hun\_Latn &  64.2 & \textbf{67.5} &hus\_Latn &  38.0 & \textbf{42.3} &hye\_Armn &  70.0 & \textbf{70.9}\\
iba\_Latn &  57.6 & \textbf{61.3} &ibo\_Latn &  \textbf{58.2} & 56.8 &ifa\_Latn &  49.1 & \textbf{55.0} &ifb\_Latn &  50.3 & \textbf{50.7}\\
ikk\_Latn &  47.8 & \textbf{51.8} &ilo\_Latn &  53.2 & \textbf{60.5} &ind\_Latn &  76.4 & \textbf{78.0} &isl\_Latn &  50.4 & \textbf{59.3}\\
ita\_Latn &  64.8 & \textbf{66.3} &ium\_Latn &  57.5 & \textbf{58.9} &ixl\_Latn &  32.2 & \textbf{38.3} &izz\_Latn &  42.4 & \textbf{49.4}\\
jam\_Latn &  64.2 & \textbf{68.5} &jav\_Latn &  45.9 & \textbf{50.5} &jpn\_Jpan &  \textbf{64.9} & 63.1 &kaa\_Cyrl &  59.3 & \textbf{66.8}\\
kab\_Latn &  23.0 & \textbf{29.9} &kac\_Latn &  \textbf{49.0} & 45.6 &kal\_Latn &  32.1 & \textbf{37.2} &kan\_Knda &  \textbf{67.1} & 65.2\\
kat\_Geor &  \textbf{60.0} & 57.2 &kaz\_Cyrl &  \textbf{65.2} & 62.8 &kbp\_Latn &  35.3 & \textbf{38.0} &kek\_Latn &  45.5 & \textbf{47.4}\\
khm\_Khmr &  \textbf{69.0} & 66.5 &kia\_Latn &  41.3 & \textbf{52.7} &kik\_Latn &  42.7 & \textbf{46.4} &kin\_Latn &  44.7 & \textbf{56.7}\\
kir\_Cyrl &  67.4 & \textbf{67.9} &kjb\_Latn &  46.8 & \textbf{48.5} &kjh\_Cyrl &  50.9 & \textbf{55.8} &kmm\_Latn &  46.2 & \textbf{57.4}\\
kmr\_Cyrl &  49.9 & \textbf{51.7} &knv\_Latn &  44.5 & \textbf{45.8} &kor\_Hang &  70.2 & \textbf{71.6} &kpg\_Latn &  64.4 & \textbf{65.4}\\
krc\_Cyrl &  57.6 & \textbf{61.8} &kri\_Latn &  59.4 & \textbf{62.8} &ksd\_Latn &  \textbf{54.8} & 52.9 &kss\_Latn &  \textbf{23.7} & 18.9\\
ksw\_Mymr &  49.0 & \textbf{50.8} &kua\_Latn &  42.6 & \textbf{45.4} &lam\_Latn &  33.3 & \textbf{38.1} &lao\_Laoo &  \textbf{72.7} & 70.7\\
lat\_Latn &  58.5 & \textbf{63.0} &lav\_Latn &  62.8 & \textbf{63.8} &ldi\_Latn &  27.6 & \textbf{35.8} &leh\_Latn &  45.1 & \textbf{46.9}\\
lhu\_Latn &  24.0 & \textbf{25.9} &lin\_Latn &  47.6 & \textbf{54.7} &lit\_Latn &  61.2 & \textbf{61.7} &loz\_Latn &  51.2 & \textbf{52.6}\\
ltz\_Latn &  \textbf{53.3} & 51.9 &lug\_Latn &  44.0 & \textbf{52.6} &luo\_Latn &  38.7 & \textbf{43.5} &lus\_Latn &  48.1 & \textbf{53.9}\\
lzh\_Hani &  61.5 & \textbf{67.2} &mad\_Latn &  60.6 & \textbf{62.5} &mah\_Latn &  34.3 & \textbf{45.3} &mai\_Deva &  \textbf{65.1} & 64.8\\
mal\_Mlym &  \textbf{7.2} & 6.1 &mam\_Latn &  29.2 & \textbf{34.9} &mar\_Deva &  62.4 & \textbf{62.6} &mau\_Latn &  \textbf{7.0} & 5.7\\
mbb\_Latn &  52.0 & \textbf{54.4} &mck\_Latn &  41.1 & \textbf{46.0} &mcn\_Latn &  36.6 & \textbf{42.8} &mco\_Latn &  24.4 & \textbf{26.8}\\
mdy\_Ethi &  49.1 & \textbf{54.4} &meu\_Latn &  49.4 & \textbf{57.8} &mfe\_Latn &  \textbf{68.8} & 68.5 &mgh\_Latn &  32.9 & \textbf{34.8}\\
mgr\_Latn &  47.3 & \textbf{50.9} &mhr\_Cyrl &  \textbf{41.6} & 40.9 &min\_Latn &  51.2 & \textbf{53.2} &miq\_Latn &  52.1 & \textbf{52.7}\\
mkd\_Cyrl &  68.8 & \textbf{71.7} &mlg\_Latn &  48.3 & \textbf{51.9} &mlt\_Latn &  51.3 & \textbf{53.2} &mos\_Latn &  36.4 & \textbf{44.4}\\
    \bottomrule
    \end{tabular}
}
    \caption{F1 scores of models on \textbf{Taxi1500} (Part I).}\label{tab:taxi_table1}
\end{table*}

\begin{table*}
\centering
\setlength{\tabcolsep}{0.7mm}{}
\resizebox{\textwidth}{!}{
    \begin{tabular}{lrr|lrr|lrr|lrr}
    \toprule
    Language & Baseline & \frameworkname &     Language & Baseline & \frameworkname &     Language & Baseline & \frameworkname &     Language & Baseline & \frameworkname \\
    \midrule
mps\_Latn &  51.6 & \textbf{56.2} &mri\_Latn &  43.2 & \textbf{49.8} &mrw\_Latn &  \textbf{48.7} & 47.9 &msa\_Latn &  46.7 & \textbf{48.9}\\
mwm\_Latn &  52.5 & \textbf{58.5} &mxv\_Latn &  16.0 & \textbf{27.6} &mya\_Mymr &  57.2 & \textbf{57.8} &myv\_Cyrl &  42.9 & \textbf{48.1}\\
mzh\_Latn &  39.2 & \textbf{41.9} &nan\_Latn &  26.2 & \textbf{33.7} &naq\_Latn &  40.3 & \textbf{45.5} &nav\_Latn &  22.2 & \textbf{25.7}\\
nbl\_Latn &  46.7 & \textbf{52.9} &nch\_Latn &  41.8 & \textbf{46.0} &ncj\_Latn &  36.2 & \textbf{42.2} &ndc\_Latn &  39.3 & \textbf{44.6}\\
nde\_Latn &  46.7 & \textbf{52.9} &ndo\_Latn &  47.3 & \textbf{51.2} &nds\_Latn &  39.1 & \textbf{48.6} &nep\_Deva &  70.7 & \textbf{72.5}\\
ngu\_Latn &  41.5 & \textbf{44.1} &nld\_Latn &  \textbf{62.7} & 62.1 &nmf\_Latn &  43.1 & \textbf{47.8} &nnb\_Latn &  38.5 & \textbf{46.0}\\
nno\_Latn &  63.1 & \textbf{65.5} &nob\_Latn &  \textbf{60.6} & 60.0 &nor\_Latn &  \textbf{61.7} & 60.8 &npi\_Deva &  69.7 & \textbf{70.1}\\
nse\_Latn &  43.9 & \textbf{45.1} &nso\_Latn &  \textbf{53.6} & 52.4 &nya\_Latn &  55.4 & \textbf{61.9} &nyn\_Latn &  43.6 & \textbf{47.0}\\
nyy\_Latn &  31.1 & \textbf{37.7} &nzi\_Latn &  34.4 & \textbf{38.3} &ori\_Orya &  \textbf{70.3} & 69.5 &ory\_Orya &  \textbf{71.4} & 69.3\\
oss\_Cyrl &  48.0 & \textbf{57.6} &ote\_Latn &  35.6 & \textbf{35.7} &pag\_Latn &  52.0 & \textbf{54.3} &pam\_Latn &  40.1 & \textbf{45.4}\\
pan\_Guru &  \textbf{67.8} & 66.4 &pap\_Latn &  65.5 & \textbf{66.2} &pau\_Latn &  42.0 & \textbf{43.3} &pcm\_Latn &  63.8 & \textbf{67.1}\\
pdt\_Latn &  58.1 & \textbf{58.7} &pes\_Arab &  \textbf{70.5} & 69.6 &pis\_Latn &  67.5 & \textbf{67.9} &pls\_Latn &  46.5 & \textbf{49.1}\\
plt\_Latn &  \textbf{52.6} & 50.1 &poh\_Latn &  47.2 & \textbf{48.2} &pol\_Latn &  64.8 & \textbf{68.8} &pon\_Latn &  53.1 & \textbf{53.9}\\
por\_Latn &  68.0 & \textbf{72.3} &prk\_Latn &  \textbf{56.4} & \textbf{56.4} &prs\_Arab &  \textbf{68.9} & \textbf{68.9} &pxm\_Latn &  \textbf{40.6} & 40.2\\
qub\_Latn &  58.3 & \textbf{59.1} &quc\_Latn &  51.0 & \textbf{53.3} &qug\_Latn &  64.5 & \textbf{68.2} &quh\_Latn &  62.4 & \textbf{68.6}\\
quw\_Latn &  53.5 & \textbf{55.6} &quy\_Latn &  \textbf{71.6} & 70.6 &quz\_Latn &  64.8 & \textbf{67.6} &qvi\_Latn &  63.2 & \textbf{65.2}\\
rap\_Latn &  47.9 & \textbf{49.4} &rar\_Latn &  45.8 & \textbf{52.7} &rmy\_Latn &  45.0 & \textbf{47.9} &ron\_Latn &  59.4 & \textbf{66.7}\\
rop\_Latn &  57.0 & \textbf{58.1} &rug\_Latn &  51.2 & \textbf{55.1} &run\_Latn &  48.2 & \textbf{53.6} &rus\_Cyrl &  69.4 & \textbf{72.3}\\
sag\_Latn &  44.7 & \textbf{47.1} &sah\_Cyrl &  59.0 & \textbf{61.6} &sba\_Latn &  38.7 & \textbf{41.0} &seh\_Latn &  47.5 & \textbf{49.9}\\
sin\_Sinh &  \textbf{67.5} & 66.6 &slk\_Latn &  60.3 & \textbf{60.6} &slv\_Latn &  62.4 & \textbf{62.6} &sme\_Latn &  36.8 & \textbf{48.8}\\
smo\_Latn &  56.2 & \textbf{61.6} &sna\_Latn &  40.9 & \textbf{45.4} &snd\_Arab &  67.7 & \textbf{68.7} &som\_Latn &  33.6 & \textbf{35.8}\\
sop\_Latn &  34.1 & \textbf{39.0} &sot\_Latn &  45.2 & \textbf{48.6} &spa\_Latn &  64.2 & \textbf{67.5} &sqi\_Latn &  \textbf{70.8} & 70.3\\
srm\_Latn &  48.7 & \textbf{52.4} &srn\_Latn &  64.0 & \textbf{64.4} &srp\_Latn &  64.9 & \textbf{69.0} &ssw\_Latn &  38.0 & \textbf{46.9}\\
sun\_Latn &  54.0 & \textbf{56.6} &suz\_Deva &  58.2 & \textbf{60.3} &swe\_Latn &  67.8 & \textbf{69.0} &swh\_Latn &  62.1 & \textbf{64.2}\\
sxn\_Latn &  49.0 & \textbf{51.8} &tam\_Taml &  72.6 & \textbf{73.7} &tat\_Cyrl &  65.5 & \textbf{68.0} &tbz\_Latn &  36.0 & \textbf{42.7}\\
tca\_Latn &  42.2 & \textbf{48.3} &tdt\_Latn &  58.2 & \textbf{66.0} &tel\_Telu &  70.9 & \textbf{71.4} &teo\_Latn &  24.0 & \textbf{26.6}\\
tgk\_Cyrl &  64.0 & \textbf{65.7} &tgl\_Latn &  57.5 & \textbf{58.8} &tha\_Thai &  65.3 & \textbf{65.5} &tih\_Latn &  56.6 & \textbf{60.3}\\
tir\_Ethi &  51.9 & \textbf{52.5} &tlh\_Latn &  64.0 & \textbf{65.3} &tob\_Latn &  \textbf{44.3} & 44.0 &toh\_Latn &  38.1 & \textbf{40.3}\\
toi\_Latn &  39.5 & \textbf{49.4} &toj\_Latn &  36.7 & \textbf{39.2} &ton\_Latn &  48.2 & \textbf{50.9} &top\_Latn &  22.9 & \textbf{26.6}\\
tpi\_Latn &  68.4 & \textbf{69.7} &tpm\_Latn &  45.8 & \textbf{51.4} &tsn\_Latn &  45.6 & \textbf{46.5} &tsz\_Latn &  37.3 & \textbf{42.8}\\
tuc\_Latn &  56.3 & \textbf{62.2} &tui\_Latn &  45.6 & \textbf{48.1} &tuk\_Latn &  56.9 & \textbf{63.5} &tum\_Latn &  48.1 & \textbf{50.1}\\
tur\_Latn &  62.6 & \textbf{65.8} &twi\_Latn &  42.1 & \textbf{47.5} &tyv\_Cyrl &  58.2 & \textbf{63.0} &tzh\_Latn &  38.8 & \textbf{44.2}\\
tzo\_Latn &  38.3 & \textbf{42.5} &udm\_Cyrl &  53.8 & \textbf{54.2} &ukr\_Cyrl &  65.3 & \textbf{67.5} &urd\_Arab &  \textbf{61.3} & 60.4\\
uzb\_Latn &  \textbf{59.7} & 58.0 &uzn\_Cyrl &  65.3 & \textbf{65.8} &ven\_Latn &  43.4 & \textbf{46.3} &vie\_Latn &  \textbf{70.0} & 69.3\\
wal\_Latn &  42.1 & \textbf{49.0} &war\_Latn &  45.6 & \textbf{52.4} &wbm\_Latn &  \textbf{57.5} & 56.1 &wol\_Latn &  34.0 & \textbf{40.4}\\
xav\_Latn &  29.9 & \textbf{33.1} &xho\_Latn &  45.3 & \textbf{48.5} &yan\_Latn &  51.2 & \textbf{53.6} &yao\_Latn &  40.0 & \textbf{46.5}\\
yap\_Latn &  39.3 & \textbf{42.0} &yom\_Latn &  36.2 & \textbf{37.8} &yor\_Latn &  \textbf{47.4} & 46.9 &yua\_Latn &  36.7 & \textbf{39.8}\\
yue\_Hani &  58.4 & \textbf{60.4} &zai\_Latn &  39.3 & \textbf{44.2} &zho\_Hani &  64.6 & \textbf{67.0} &zlm\_Latn &  \textbf{70.3} & 69.7\\
zom\_Latn &  47.2 & \textbf{49.9} &zsm\_Latn &  \textbf{69.7} & 68.0 &zul\_Latn &  50.6 & \textbf{53.4} & &\\
    \bottomrule
    \end{tabular}
}
    \caption{F1 scores of models on \textbf{Taxi1500} (Part II).}\label{tab:taxi_table2}
\end{table*}

%% file: sib200.tex
\begin{table*}
\centering
\setlength{\tabcolsep}{0.7mm}{}
\resizebox{\textwidth}{!}{
    \begin{tabular}{lrr|lrr|lrr|lrr}
    \toprule
    Language & Baseline & \frameworkname &     Language & Baseline & \frameworkname &     Language & Baseline & \frameworkname &     Language & Baseline & \frameworkname \\
    \midrule
    ace\_Latn &  71.5 & \textbf{73.6} &acm\_Arab &  82.2 & \textbf{83.0} &afr\_Latn &  82.3 & \textbf{82.7} &ajp\_Arab &  \textbf{83.4} & 81.8\\
aka\_Latn &  62.2 & \textbf{67.2} &als\_Latn &  82.4 & \textbf{84.4} &amh\_Ethi &  \textbf{74.2} & 73.6 &apc\_Arab &  \textbf{83.9} & 82.9\\
arb\_Arab &  \textbf{83.8} & 82.9 &ary\_Arab &  \textbf{81.5} & 80.2 &arz\_Arab &  \textbf{84.5} & 84.1 &asm\_Beng &  83.6 & \textbf{84.2}\\
ast\_Latn &  \textbf{88.4} & 88.0 &ayr\_Latn &  51.1 & \textbf{53.8} &azb\_Arab &  71.5 & \textbf{74.7} &azj\_Latn &  87.0 & \textbf{88.0}\\
bak\_Cyrl &  84.6 & \textbf{86.6} &bam\_Latn &  \textbf{47.9} & 47.6 &ban\_Latn &  80.3 & \textbf{83.0} &bel\_Cyrl &  \textbf{83.7} & 83.4\\
bem\_Latn &  63.0 & \textbf{63.9} &ben\_Beng &  83.3 & \textbf{84.3} &bjn\_Latn &  77.1 & \textbf{78.5} &bod\_Tibt &  \textbf{73.5} & 69.2\\
bos\_Latn &  86.5 & \textbf{88.2} &bul\_Cyrl &  86.1 & \textbf{87.5} &cat\_Latn &  84.8 & \textbf{86.4} &ceb\_Latn &  81.8 & \textbf{84.6}\\
ces\_Latn &  \textbf{89.1} & 86.9 &cjk\_Latn &  46.6 & \textbf{48.1} &ckb\_Arab &  \textbf{83.9} & 80.2 &crh\_Latn &  74.0 & \textbf{76.2}\\
cym\_Latn &  \textbf{75.9} & 75.4 &dan\_Latn &  86.8 & \textbf{87.4} &deu\_Latn &  86.5 & \textbf{87.8} &dyu\_Latn &  42.6 & \textbf{44.5}\\
dzo\_Tibt &  68.7 & \textbf{72.6} &ell\_Grek &  79.5 & \textbf{80.0} &eng\_Latn &  \textbf{90.8} & 90.0 &epo\_Latn &  \textbf{83.8} & 82.2\\
est\_Latn &  80.6 & \textbf{81.6} &eus\_Latn &  82.1 & \textbf{82.2} &ewe\_Latn &  49.3 & \textbf{51.5} &fao\_Latn &  83.7 & \textbf{84.9}\\
fij\_Latn &  56.1 & \textbf{58.0} &fin\_Latn &  82.1 & \textbf{82.9} &fon\_Latn &  41.7 & \textbf{44.6} &fra\_Latn &  87.9 & \textbf{89.6}\\
fur\_Latn &  77.6 & \textbf{80.2} &gla\_Latn &  \textbf{57.6} & 54.3 &gle\_Latn &  62.2 & \textbf{64.1} &glg\_Latn &  87.8 & \textbf{89.0}\\
grn\_Latn &  \textbf{75.0} & 74.5 &guj\_Gujr &  83.9 & \textbf{84.7} &hat\_Latn &  77.4 & \textbf{79.1} &hau\_Latn &  \textbf{62.7} & 62.1\\
heb\_Hebr &  77.9 & \textbf{79.2} &hin\_Deva &  84.1 & \textbf{84.4} &hne\_Deva &  77.9 & \textbf{80.1} &hrv\_Latn &  87.3 & \textbf{89.0}\\
hun\_Latn &  86.8 & \textbf{87.6} &hye\_Armn &  \textbf{83.0} & 82.5 &ibo\_Latn &  72.3 & \textbf{74.1} &ilo\_Latn &  75.8 & \textbf{79.6}\\
ind\_Latn &  88.7 & \textbf{89.1} &isl\_Latn &  78.5 & \textbf{79.1} &ita\_Latn &  87.7 & \textbf{89.2} &jav\_Latn &  80.2 & \textbf{80.3}\\
jpn\_Jpan &  87.1 & \textbf{87.9} &kab\_Latn &  31.1 & \textbf{36.9} &kac\_Latn &  49.3 & \textbf{52.3} &kam\_Latn &  49.1 & \textbf{49.5}\\
kan\_Knda &  \textbf{83.2} & 82.0 &kat\_Geor &  81.8 & \textbf{83.7} &kaz\_Cyrl &  84.2 & \textbf{84.9} &kbp\_Latn &  \textbf{45.1} & 44.2\\
kea\_Latn &  75.4 & \textbf{77.0} &khm\_Khmr &  84.3 & \textbf{84.4} &kik\_Latn &  57.1 & \textbf{59.9} &kin\_Latn &  69.5 & \textbf{70.5}\\
kir\_Cyrl &  \textbf{80.7} & 80.3 &kmb\_Latn &  48.2 & \textbf{49.5} &kmr\_Latn &  \textbf{70.7} & 70.0 &kon\_Latn &  65.3 & \textbf{69.2}\\
kor\_Hang &  \textbf{85.2} & 83.9 &lao\_Laoo &  \textbf{85.1} & 84.2 &lij\_Latn &  77.7 & \textbf{79.6} &lim\_Latn &  74.7 & \textbf{75.2}\\
lin\_Latn &  69.3 & \textbf{71.4} &lit\_Latn &  \textbf{86.5} & 84.7 &lmo\_Latn &  77.7 & \textbf{79.1} &ltz\_Latn &  76.6 & \textbf{79.1}\\
lua\_Latn &  \textbf{59.1} & 56.4 &lug\_Latn &  55.5 & \textbf{59.1} &luo\_Latn &  52.6 & \textbf{53.0} &lus\_Latn &  65.3 & \textbf{67.9}\\
lvs\_Latn &  \textbf{84.4} & 83.6 &mai\_Deva &  83.4 & \textbf{84.0} &mal\_Mlym &  \textbf{80.6} & 79.9 &mar\_Deva &  \textbf{84.1} & 82.5\\
min\_Latn &  77.7 & \textbf{79.6} &mkd\_Cyrl &  83.3 & \textbf{84.6} &mlt\_Latn &  82.9 & \textbf{83.0} &mos\_Latn &  44.9 & \textbf{46.6}\\
mri\_Latn &  54.4 & \textbf{59.3} &mya\_Mymr &  80.1 & \textbf{81.6} &nld\_Latn &  \textbf{86.5} & 85.8 &nno\_Latn &  \textbf{86.6} & 86.4\\
nob\_Latn &  85.8 & \textbf{86.1} &npi\_Deva &  \textbf{86.8} & 86.0 &nso\_Latn &  61.3 & \textbf{61.9} &nya\_Latn &  71.1 & \textbf{72.7}\\
oci\_Latn &  83.1 & \textbf{84.9} &ory\_Orya &  79.7 & \textbf{80.3} &pag\_Latn &  78.7 & \textbf{79.7} &pan\_Guru &  77.4 & \textbf{79.0}\\
pap\_Latn &  77.2 & \textbf{79.0} &pes\_Arab &  87.6 & \textbf{89.2} &plt\_Latn &  68.4 & \textbf{68.5} &pol\_Latn &  86.4 & \textbf{86.7}\\
por\_Latn &  87.3 & \textbf{88.6} &prs\_Arab &  85.8 & \textbf{88.4} &quy\_Latn &  63.7 & \textbf{64.0} &ron\_Latn &  \textbf{86.4} & 84.5\\
run\_Latn &  \textbf{68.3} & 67.2 &rus\_Cyrl &  87.6 & \textbf{87.9} &sag\_Latn &  52.4 & \textbf{55.1} &san\_Deva &  \textbf{77.9} & 77.8\\
sat\_Olck &  53.0 & \textbf{57.4} &scn\_Latn &  77.6 & \textbf{78.2} &sin\_Sinh &  \textbf{84.5} & 84.1 &slk\_Latn &  86.1 & \textbf{87.0}\\
slv\_Latn &  \textbf{86.4} & 85.5 &smo\_Latn &  73.4 & \textbf{74.1} &sna\_Latn &  \textbf{59.3} & 58.0 &snd\_Arab &  72.1 & \textbf{76.9}\\
som\_Latn &  \textbf{61.8} & 59.8 &sot\_Latn &  65.3 & \textbf{67.6} &spa\_Latn &  \textbf{86.4} & 86.2 &srd\_Latn &  74.0 & \textbf{75.8}\\
srp\_Cyrl &  \textbf{85.8} & 85.2 &ssw\_Latn &  67.5 & \textbf{68.1} &sun\_Latn &  84.0 & \textbf{85.2} &swe\_Latn &  86.6 & \textbf{87.3}\\
swh\_Latn &  76.0 & \textbf{78.6} &szl\_Latn &  74.3 & \textbf{75.5} &tam\_Taml &  80.6 & \textbf{84.3} &tat\_Cyrl &  84.0 & \textbf{85.2}\\
tel\_Telu &  85.3 & \textbf{85.7} &tgk\_Cyrl &  \textbf{81.6} & 80.9 &tgl\_Latn &  81.9 & \textbf{83.0} &tha\_Thai &  87.4 & \textbf{88.9}\\
tir\_Ethi &  59.9 & \textbf{61.4} &tpi\_Latn &  80.6 & \textbf{82.3} &tsn\_Latn &  \textbf{59.1} & 55.2 &tso\_Latn &  59.3 & \textbf{61.2}\\
tuk\_Latn &  \textbf{78.3} & 78.2 &tum\_Latn &  70.3 & \textbf{70.8} &tur\_Latn &  82.9 & \textbf{83.6} &twi\_Latn &  61.4 & \textbf{68.0}\\
uig\_Arab &  77.7 & \textbf{80.0} &ukr\_Cyrl &  \textbf{84.7} & 84.5 &umb\_Latn &  \textbf{45.9} & 45.8 &urd\_Arab &  81.3 & \textbf{81.9}\\
vec\_Latn &  \textbf{82.0} & 81.1 &vie\_Latn &  84.9 & \textbf{85.8} &war\_Latn &  81.7 & \textbf{83.4} &wol\_Latn &  49.2 & \textbf{52.1}\\
xho\_Latn &  62.4 & \textbf{64.0} &yor\_Latn &  46.6 & \textbf{51.8} &zsm\_Latn &  \textbf{87.2} & 86.6 &zul\_Latn &  \textbf{73.8} & 73.6\\
    \bottomrule
    \end{tabular}
}
    \caption{F1 scores of models on \textbf{SIB200}.}\label{tab:sib200_table1}
\end{table*}

%% file: ner.tex
\begin{table*}
\centering
\setlength{\tabcolsep}{0.7mm}{}
\resizebox{\textwidth}{!}{
    \begin{tabular}{lrr|lrr|lrr|lrr}
    \toprule
    Language & Baseline & \frameworkname &     Language & Baseline & \frameworkname &     Language & Baseline & \frameworkname &     Language & Baseline & \frameworkname \\
    \midrule
ace\_Latn &  \textbf{42.8} & 42.6 &afr\_Latn &  76.4 & \textbf{76.8} &als\_Latn &  81.7 & \textbf{82.2} &amh\_Ethi &  \textbf{40.8} & 39.0\\
ara\_Arab &  55.9 & \textbf{56.5} &arg\_Latn &  78.9 & \textbf{80.0} &arz\_Arab &  56.8 & \textbf{58.4} &asm\_Beng &  \textbf{66.0} & 64.8\\
ast\_Latn &  82.3 & \textbf{84.5} &aym\_Latn &  \textbf{45.9} & 43.8 &aze\_Latn &  65.4 & \textbf{65.9} &bak\_Cyrl &  \textbf{61.3} & 60.2\\
bar\_Latn &  68.2 & \textbf{70.0} &bel\_Cyrl &  74.4 & \textbf{74.6} &ben\_Beng &  \textbf{71.1} & 70.5 &bih\_Deva &  \textbf{55.9} & 55.2\\
bod\_Tibt &  27.1 & \textbf{35.0} &bos\_Latn &  \textbf{73.0} & 72.6 &bre\_Latn &  \textbf{64.6} & 63.9 &bul\_Cyrl &  \textbf{75.6} & 75.2\\
cat\_Latn &  \textbf{83.8} & \textbf{83.8} &cbk\_Latn &  \textbf{53.9} & 51.6 &ceb\_Latn &  54.0 & \textbf{57.5} &ces\_Latn &  78.6 & \textbf{78.7}\\
che\_Cyrl &  45.6 & \textbf{55.7} &chv\_Cyrl &  \textbf{77.4} & 75.2 &ckb\_Arab &  73.2 & \textbf{73.3} &cos\_Latn &  59.1 & \textbf{59.7}\\
crh\_Latn &  \textbf{51.9} & 50.2 &csb\_Latn &  61.9 & \textbf{62.0} &cym\_Latn &  \textbf{62.2} & 60.2 &dan\_Latn &  \textbf{81.7} & \textbf{81.7}\\
deu\_Latn &  76.0 & \textbf{76.7} &diq\_Latn &  51.7 & \textbf{53.4} &div\_Thaa &  48.2 & \textbf{55.2} &ell\_Grek &  \textbf{73.0} & 72.9\\
eml\_Latn &  42.0 & \textbf{42.1} &eng\_Latn &  \textbf{83.5} & 83.4 &epo\_Latn &  68.1 & \textbf{69.0} &est\_Latn &  72.9 & \textbf{73.6}\\
eus\_Latn &  57.5 & \textbf{59.1} &ext\_Latn &  45.0 & \textbf{46.2} &fao\_Latn &  \textbf{70.9} & 69.6 &fas\_Arab &  \textbf{52.7} & 51.5\\
fin\_Latn &  75.0 & \textbf{75.4} &fra\_Latn &  76.4 & \textbf{76.8} &frr\_Latn &  \textbf{55.3} & 54.5 &fry\_Latn &  \textbf{77.2} & 76.3\\
fur\_Latn &  \textbf{57.5} & 57.1 &gla\_Latn &  59.5 & \textbf{65.4} &gle\_Latn &  72.7 & \textbf{72.8} &glg\_Latn &  79.9 & \textbf{80.4}\\
grn\_Latn &  54.0 & \textbf{54.1} &guj\_Gujr &  58.8 & \textbf{59.2} &hbs\_Latn &  62.7 & \textbf{65.5} &heb\_Hebr &  50.4 & \textbf{51.7}\\
hin\_Deva &  68.4 & \textbf{68.9} &hrv\_Latn &  77.1 & \textbf{77.8} &hsb\_Latn &  74.2 & \textbf{75.9} &hun\_Latn &  76.2 & \textbf{76.7}\\
hye\_Armn &  54.1 & \textbf{57.2} &ibo\_Latn &  \textbf{57.3} & \textbf{57.3} &ido\_Latn &  79.0 & \textbf{79.9} &ilo\_Latn &  73.6 & \textbf{76.8}\\
ina\_Latn &  57.8 & \textbf{58.5} &ind\_Latn &  62.2 & \textbf{64.0} &isl\_Latn &  \textbf{72.4} & 72.1 &ita\_Latn &  78.4 & \textbf{78.6}\\
jav\_Latn &  \textbf{58.9} & 57.8 &jbo\_Latn &  25.0 & \textbf{25.1} &jpn\_Jpan &  \textbf{21.0} & 17.5 &kan\_Knda &  58.5 & \textbf{58.7}\\
kat\_Geor &  67.5 & \textbf{68.2} &kaz\_Cyrl &  50.0 & \textbf{51.1} &khm\_Khmr &  42.4 & \textbf{43.1} &kin\_Latn &  63.5 & \textbf{69.0}\\
kir\_Cyrl &  \textbf{45.4} & 44.2 &kor\_Hang &  \textbf{52.7} & 51.2 &ksh\_Latn &  58.0 & \textbf{59.2} &kur\_Latn &  63.1 & \textbf{64.8}\\
lat\_Latn &  \textbf{75.7} & 74.7 &lav\_Latn &  72.4 & \textbf{74.9} &lij\_Latn &  42.7 & \textbf{46.3} &lim\_Latn &  70.3 & \textbf{71.0}\\
lin\_Latn &  \textbf{51.5} & 49.2 &lit\_Latn &  \textbf{75.0} & 74.6 &lmo\_Latn &  \textbf{75.3} & 74.1 &ltz\_Latn &  68.6 & \textbf{69.0}\\
lzh\_Hani &  \textbf{14.6} & 11.5 &mal\_Mlym &  \textbf{62.8} & 61.7 &mar\_Deva &  \textbf{63.0} & 62.0 &mhr\_Cyrl &  60.6 & \textbf{61.9}\\
min\_Latn &  41.5 & \textbf{42.9} &mkd\_Cyrl &  75.7 & \textbf{76.3} &mlg\_Latn &  57.8 & \textbf{58.7} &mlt\_Latn &  66.9 & \textbf{71.3}\\
mon\_Cyrl &  66.0 & \textbf{68.9} &mri\_Latn &  \textbf{49.5} & 48.0 &msa\_Latn &  68.7 & \textbf{69.0} &mwl\_Latn &  48.9 & \textbf{51.1}\\
mya\_Mymr &  \textbf{53.8} & 53.6 &mzn\_Arab &  44.9 & \textbf{47.4} &nan\_Latn &  83.6 & \textbf{85.0} &nap\_Latn &  59.0 & \textbf{59.3}\\
nds\_Latn &  \textbf{77.8} & 77.0 &nep\_Deva &  \textbf{63.1} & 61.7 &nld\_Latn &  \textbf{81.4} & 81.1 &nno\_Latn &  76.5 & \textbf{77.3}\\
nor\_Latn &  76.6 & \textbf{77.7} &oci\_Latn &  68.6 & \textbf{70.7} &ori\_Orya &  31.7 & \textbf{31.9} &oss\_Cyrl &  \textbf{53.5} & 51.3\\
pan\_Guru &  \textbf{49.7} & 48.1 &pms\_Latn &  78.9 & \textbf{80.3} &pnb\_Arab &  \textbf{65.1} & 64.7 &pol\_Latn &  78.0 & \textbf{78.1}\\
por\_Latn &  77.6 & \textbf{78.7} &pus\_Arab &  40.5 & \textbf{42.0} &que\_Latn &  66.1 & \textbf{67.4} &roh\_Latn &  \textbf{62.0} & 58.8\\
ron\_Latn &  \textbf{76.0} & 75.6 &rus\_Cyrl &  69.0 & \textbf{69.1} &sah\_Cyrl &  \textbf{75.3} & 69.3 &san\_Deva &  35.5 & \textbf{37.1}\\
scn\_Latn &  65.4 & \textbf{66.0} &sco\_Latn &  84.5 & \textbf{87.3} &sgs\_Latn &  60.7 & \textbf{64.0} &sin\_Sinh &  \textbf{53.0} & 49.4\\
slk\_Latn &  \textbf{77.9} & 77.3 &slv\_Latn &  \textbf{80.2} & 80.1 &snd\_Arab &  \textbf{44.3} & 41.9 &som\_Latn &  52.2 & \textbf{55.8}\\
spa\_Latn &  74.0 & \textbf{76.5} &sqi\_Latn &  \textbf{77.7} & 77.6 &srp\_Cyrl &  63.5 & \textbf{64.3} &sun\_Latn &  \textbf{55.4} & 53.8\\
swa\_Latn &  \textbf{69.1} & 69.0 &swe\_Latn &  70.3 & \textbf{73.0} &szl\_Latn &  68.1 & \textbf{70.8} &tam\_Taml &  54.9 & \textbf{55.3}\\
tat\_Cyrl &  \textbf{65.9} & 63.0 &tel\_Telu &  \textbf{50.2} & 49.1 &tgk\_Cyrl &  63.0 & \textbf{66.0} &tgl\_Latn &  75.5 & \textbf{76.8}\\
tha\_Thai &  \textbf{4.6} & 2.2 &tuk\_Latn &  56.3 & \textbf{56.5} &tur\_Latn &  76.1 & \textbf{76.6} &uig\_Arab &  47.8 & \textbf{48.6}\\
ukr\_Cyrl &  \textbf{76.6} & \textbf{76.6} &urd\_Arab &  \textbf{66.2} & 65.1 &uzb\_Latn &  73.1 & \textbf{74.5} &vec\_Latn &  \textbf{68.6} & 67.7\\
vep\_Latn &  71.1 & \textbf{71.2} &vie\_Latn &  73.0 & \textbf{73.1} &vls\_Latn &  75.5 & \textbf{76.5} &vol\_Latn &  59.6 & \textbf{59.7}\\
war\_Latn &  66.4 & \textbf{66.6} &wuu\_Hani &  \textbf{32.3} & 28.5 &xmf\_Geor &  63.6 & \textbf{65.7} &yid\_Hebr &  50.4 & \textbf{55.7}\\
yor\_Latn &  60.7 & \textbf{61.6} &yue\_Hani &  \textbf{23.7} & 21.6 &zea\_Latn &  \textbf{66.8} & 63.7 &zho\_Hani &  \textbf{24.7} & 20.9\\
    \bottomrule
    \end{tabular}
}
    \caption{F1 scores of models on \textbf{NER}.}\label{tab:ner_table1}
\end{table*}

%% file: pos.tex
\begin{table*}
\centering
\setlength{\tabcolsep}{0.7mm}{}
\resizebox{\textwidth}{!}{
    \begin{tabular}{lrr|lrr|lrr|lrr}
    \toprule
    Language & Baseline & \frameworkname &     Language & Baseline & \frameworkname &     Language & Baseline & \frameworkname &     Language & Baseline & \frameworkname \\
    \midrule
afr\_Latn &  87.8 & \textbf{88.2} &ajp\_Arab &  \textbf{70.4} & 69.2 &aln\_Latn &  \textbf{52.9} & 50.7 &amh\_Ethi &  65.9 & \textbf{67.0}\\
ara\_Arab &  \textbf{66.5} & \textbf{66.5} &bam\_Latn &  \textbf{42.2} & 41.5 &bel\_Cyrl &  \textbf{86.0} & 85.5 &ben\_Beng &  \textbf{83.2} & 83.0\\
bre\_Latn &  \textbf{61.0} & 60.0 &bul\_Cyrl &  88.1 & \textbf{88.2} &cat\_Latn &  86.5 & \textbf{86.7} &ceb\_Latn &  \textbf{65.2} & \textbf{65.2}\\
ces\_Latn &  \textbf{84.7} & \textbf{84.7} &cym\_Latn &  \textbf{65.7} & 64.6 &dan\_Latn &  \textbf{90.4} & \textbf{90.4} &deu\_Latn &  \textbf{87.9} & 87.6\\
ell\_Grek &  \textbf{86.1} & 84.9 &eng\_Latn &  \textbf{96.0} & \textbf{96.0} &est\_Latn &  \textbf{83.7} & \textbf{83.7} &eus\_Latn &  \textbf{65.2} & 64.2\\
fao\_Latn &  \textbf{88.8} & 88.3 &fas\_Arab &  71.3 & \textbf{71.9} &fin\_Latn &  \textbf{82.1} & 81.7 &fra\_Latn &  \textbf{86.2} & 85.8\\
gla\_Latn &  \textbf{57.9} & 57.5 &gle\_Latn &  \textbf{64.0} & \textbf{64.0} &glg\_Latn &  83.1 & \textbf{83.5} &glv\_Latn &  \textbf{51.7} & 51.0\\
grc\_Grek &  \textbf{71.7} & 71.0 &grn\_Latn &  19.7 & \textbf{21.3} &gsw\_Latn &  79.7 & \textbf{80.0} &hbo\_Hebr &  34.0 & \textbf{38.6}\\
heb\_Hebr &  \textbf{68.5} & 68.0 &hin\_Deva &  70.4 & \textbf{71.8} &hrv\_Latn &  \textbf{85.6} & 85.5 &hsb\_Latn &  \textbf{83.2} & 82.9\\
hun\_Latn &  81.5 & \textbf{82.3} &hye\_Armn &  84.0 & \textbf{84.4} &hyw\_Armn &  81.2 & \textbf{81.4} &ind\_Latn &  \textbf{83.6} & 83.3\\
isl\_Latn &  \textbf{82.9} & 82.7 &ita\_Latn &  87.8 & \textbf{88.7} &jav\_Latn &  73.4 & \textbf{74.0} &jpn\_Jpan &  22.6 & \textbf{28.9}\\
kaz\_Cyrl &  \textbf{76.5} & 76.0 &kmr\_Latn &  74.1 & \textbf{74.2} &kor\_Hang &  \textbf{52.8} & 52.2 &lat\_Latn &  \textbf{72.8} & 71.7\\
lav\_Latn &  \textbf{83.6} & \textbf{83.6} &lij\_Latn &  \textbf{76.4} & 75.6 &lit\_Latn &  \textbf{81.5} & 81.3 &lzh\_Hani &  \textbf{23.7} & 21.3\\
mal\_Mlym &  \textbf{86.4} & 86.1 &mar\_Deva &  \textbf{81.4} & 80.7 &mlt\_Latn &  79.5 & \textbf{80.0} &myv\_Cyrl &  64.4 & \textbf{64.6}\\
nap\_Latn &  70.6 & \textbf{75.9} &nds\_Latn &  77.0 & \textbf{77.8} &nld\_Latn &  88.2 & \textbf{88.3} &nor\_Latn &  \textbf{88.2} & \textbf{88.2}\\
pcm\_Latn &  \textbf{57.3} & \textbf{57.3} &pol\_Latn &  \textbf{83.5} & 83.2 &por\_Latn &  88.0 & \textbf{88.1} &quc\_Latn &  \textbf{61.1} & 58.8\\
ron\_Latn &  81.4 & \textbf{81.7} &rus\_Cyrl &  \textbf{88.9} & 88.5 &sah\_Cyrl &  \textbf{74.0} & 73.0 &san\_Deva &  \textbf{25.7} & 24.2\\
sin\_Sinh &  56.0 & \textbf{56.5} &slk\_Latn &  84.3 & \textbf{84.9} &slv\_Latn &  \textbf{77.2} & 77.0 &sme\_Latn &  73.0 & \textbf{73.1}\\
spa\_Latn &  87.6 & \textbf{87.8} &sqi\_Latn &  76.3 & \textbf{76.9} &srp\_Latn &  \textbf{85.3} & \textbf{85.3} &swe\_Latn &  92.5 & \textbf{92.6}\\
tam\_Taml &  73.6 & \textbf{73.7} &tat\_Cyrl &  70.2 & \textbf{71.1} &tel\_Telu &  \textbf{81.8} & 81.7 &tgl\_Latn &  \textbf{75.5} & 75.0\\
tha\_Thai &  55.9 & \textbf{56.7} &tur\_Latn &  \textbf{71.3} & 71.1 &uig\_Arab &  \textbf{68.2} & \textbf{68.2} &ukr\_Cyrl &  85.0 & \textbf{85.2}\\
urd\_Arab &  62.0 & \textbf{64.6} &vie\_Latn &  \textbf{68.3} & 67.4 &wol\_Latn &  \textbf{61.6} & 59.5 &xav\_Latn &  \textbf{11.8} & 10.5\\
yor\_Latn &  62.3 & \textbf{62.6} &yue\_Hani &  37.4 & \textbf{41.3} &zho\_Hani &  38.7 & \textbf{44.8} & & \\
    \bottomrule
    \end{tabular}
}
    \caption{F1 scores of models on \textbf{POS}.}\label{tab:pos_table1}
\end{table*}

%% file: taxi1500_donor.tex
\begin{table*}
\centering
\setlength{\tabcolsep}{0.7mm}{}
\resizebox{\textwidth}{!}{
    \begin{tabular}{lrr|lrr|lrr|lrr}
    \toprule
    Language & English & Closest donor &     Language & English & Closest donor &     Language & English & Closest donor &     Language & English & Closest donor \\
    \midrule
ace\_Latn & \textbf{63.3} & 60.1 & ach\_Latn & 35.6 & \textbf{48.1} & acr\_Latn & \textbf{48.8} & 46.7 & afr\_Latn & \textbf{58.6} & 58.5 \\
ahk\_Latn & 5.4 & \textbf{8.3} & aka\_Latn & \textbf{44.9} & 41.2 & aln\_Latn & \textbf{56.2} & 54.7 & als\_Latn & \textbf{57.1} & 57.1 \\
alz\_Latn & 34.1 & \textbf{43.0} & aoj\_Latn & 40.9 & \textbf{46.2} & arb\_Arab & \textbf{55.4} & 55.4 & arn\_Latn & 43.1 & \textbf{44.4} \\
arz\_Arab & 33.7 & \textbf{40.3} & asm\_Beng & 53.4 & \textbf{61.5} & ayr\_Latn & 52.7 & \textbf{62.2} & azb\_Arab & \textbf{61.0} & 61.0 \\
bak\_Cyrl & 54.7 & \textbf{59.7} & bam\_Latn & 48.9 & \textbf{55.6} & ban\_Latn & \textbf{43.0} & 42.5 & bar\_Latn & \textbf{47.8} & 43.3 \\
bci\_Latn & 34.6 & \textbf{37.1} & bcl\_Latn & 54.2 & \textbf{60.5} & bel\_Cyrl & 59.1 & \textbf{61.5} & bem\_Latn & 44.2 & \textbf{49.4} \\
bhw\_Latn & \textbf{50.2} & 46.9 & bim\_Latn & 47.3 & \textbf{55.1} & bis\_Latn & \textbf{68.4} & 68.1 & bqc\_Latn & 33.2 & \textbf{41.6} \\
btx\_Latn & \textbf{56.7} & 53.8 & bul\_Cyrl & 62.5 & \textbf{62.6} & bum\_Latn & 39.6 & \textbf{42.2} & bzj\_Latn & \textbf{65.7} & 60.3 \\
cac\_Latn & 43.8 & \textbf{46.0} & cak\_Latn & 51.0 & \textbf{57.9} & caq\_Latn & 42.7 & \textbf{51.0} & cat\_Latn & 61.2 & \textbf{62.3} \\
cce\_Latn & \textbf{43.8} & 38.0 & ceb\_Latn & \textbf{49.8} & 49.1 & ces\_Latn & 63.3 & \textbf{63.7} & cfm\_Latn & \textbf{58.3} & 57.1 \\
chk\_Latn & \textbf{42.8} & 38.9 & chv\_Cyrl & 60.3 & \textbf{64.3} & ckb\_Arab & 58.3 & \textbf{67.0} & cmn\_Hani & 60.8 & \textbf{73.0} \\
crh\_Cyrl & 61.4 & \textbf{67.7} & crs\_Latn & 62.3 & \textbf{63.5} & csy\_Latn & \textbf{58.3} & 56.7 & ctd\_Latn & \textbf{56.6} & 55.8 \\
cuk\_Latn & 39.1 & \textbf{40.8} & cym\_Latn & \textbf{51.9} & 46.0 & dan\_Latn & \textbf{58.1} & 54.0 & deu\_Latn & \textbf{51.5} & 51.5 \\
dln\_Latn & \textbf{54.4} & 54.4 & dtp\_Latn & 51.5 & \textbf{51.6} & dyu\_Latn & \textbf{55.6} & 48.2 & dzo\_Tibt & 50.6 & \textbf{58.1} \\
ell\_Grek & \textbf{56.9} & 53.9 & eng\_Latn & \textbf{78.0} & 78.0 & enm\_Latn & \textbf{70.8} & 67.0 & epo\_Latn & \textbf{58.3} & 58.3 \\
eus\_Latn & \textbf{25.2} & 21.4 & ewe\_Latn & 46.4 & \textbf{52.1} & fao\_Latn & 56.5 & \textbf{64.8} & fas\_Arab & 69.6 & \textbf{70.2} \\
fil\_Latn & 56.7 & \textbf{58.7} & fin\_Latn & \textbf{56.4} & 55.7 & fon\_Latn & \textbf{36.8} & 35.4 & fra\_Latn & \textbf{66.8} & 66.8 \\
gaa\_Latn & 36.9 & \textbf{47.7} & gil\_Latn & 40.4 & \textbf{47.2} & giz\_Latn & 48.4 & \textbf{48.5} & gkn\_Latn & \textbf{40.0} & 34.1 \\
gla\_Latn & \textbf{45.6} & 45.6 & gle\_Latn & 41.8 & \textbf{45.1} & glv\_Latn & 37.3 & \textbf{48.7} & gom\_Latn & 34.8 & \textbf{41.6} \\
guc\_Latn & \textbf{39.6} & 37.6 & gug\_Latn & 39.0 & \textbf{46.0} & guj\_Gujr & 67.1 & \textbf{70.4} & gur\_Latn & 37.0 & \textbf{44.2} \\
gya\_Latn & 39.6 & \textbf{41.8} & gym\_Latn & 45.4 & \textbf{52.9} & hat\_Latn & \textbf{63.0} & 60.0 & hau\_Latn & 54.0 & \textbf{59.6} \\
heb\_Hebr & \textbf{16.7} & 15.2 & hif\_Latn & 42.4 & \textbf{53.6} & hil\_Latn & \textbf{63.7} & 61.6 & hin\_Deva & \textbf{64.8} & 64.8 \\
hne\_Deva & 64.1 & \textbf{67.5} & hnj\_Latn & 61.5 & \textbf{63.2} & hra\_Latn & 48.2 & \textbf{53.1} & hrv\_Latn & \textbf{62.7} & 60.7 \\
hun\_Latn & 65.2 & \textbf{65.9} & hus\_Latn & 37.6 & \textbf{40.7} & hye\_Armn & 67.2 & \textbf{69.3} & iba\_Latn & 57.9 & \textbf{59.2} \\
ifa\_Latn & 49.7 & \textbf{51.5} & ifb\_Latn & \textbf{48.3} & 48.1 & ikk\_Latn & 46.6 & \textbf{52.5} & ilo\_Latn & \textbf{58.8} & 55.7 \\
isl\_Latn & 53.5 & \textbf{61.2} & ita\_Latn & 62.8 & \textbf{67.1} & ium\_Latn & 51.4 & \textbf{58.0} & ixl\_Latn & 36.6 & \textbf{38.2} \\
jam\_Latn & \textbf{66.1} & 61.0 & jav\_Latn & 43.9 & \textbf{47.6} & jpn\_Jpan & \textbf{58.6} & 58.6 & kaa\_Latn & 57.7 & \textbf{62.6} \\
kac\_Latn & 44.5 & \textbf{47.3} & kal\_Latn & 31.5 & \textbf{34.5} & kan\_Knda & 60.6 & \textbf{67.5} & kat\_Geor & 55.2 & \textbf{62.2} \\
kbp\_Latn & 34.9 & \textbf{39.5} & kek\_Latn & \textbf{41.5} & 40.3 & khm\_Khmr & \textbf{64.7} & 64.7 & kia\_Latn & 48.0 & \textbf{51.7} \\
kin\_Latn & 47.2 & \textbf{52.5} & kir\_Cyrl & 61.1 & \textbf{64.7} & kjb\_Latn & 44.7 & \textbf{48.1} & kjh\_Cyrl & \textbf{52.3} & 51.1 \\
kmr\_Cyrl & 45.5 & \textbf{53.1} & knv\_Latn & \textbf{42.6} & 40.5 & kor\_Hang & 69.8 & \textbf{71.3} & kpg\_Latn & \textbf{64.1} & 57.4 \\
kri\_Latn & \textbf{63.2} & 56.0 & ksd\_Latn & 54.2 & \textbf{54.4} & kss\_Latn & 16.2 & \textbf{21.6} & ksw\_Mymr & \textbf{50.4} & 50.3 \\
lam\_Latn & 34.7 & \textbf{35.6} & lao\_Laoo & 69.1 & \textbf{72.7} & lat\_Latn & 57.2 & \textbf{62.9} & lav\_Latn & \textbf{60.4} & 57.7 \\
leh\_Latn & \textbf{43.5} & 37.2 & lhu\_Latn & 22.3 & \textbf{29.0} & lin\_Latn & 47.1 & \textbf{54.7} & lit\_Latn & 58.3 & \textbf{59.7} \\
ltz\_Latn & \textbf{48.2} & 48.2 & lug\_Latn & \textbf{46.1} & 39.0 & luo\_Latn & 40.6 & \textbf{41.2} & lus\_Latn & \textbf{51.6} & 51.6 \\
mad\_Latn & 55.3 & \textbf{63.0} & mah\_Latn & \textbf{41.6} & 38.3 & mai\_Deva & \textbf{62.7} & 60.5 & mam\_Latn & \textbf{33.9} & 33.2 \\
mau\_Latn & 5.5 & \textbf{8.4} & mbb\_Latn & 52.6 & \textbf{53.1} & mck\_Latn & \textbf{41.9} & 41.2 & mcn\_Latn & 37.7 & \textbf{39.3} \\
mdy\_Ethi & 51.6 & \textbf{57.6} & meu\_Latn & 54.9 & \textbf{55.8} & mfe\_Latn & 66.0 & \textbf{66.2} & mgh\_Latn & 30.3 & \textbf{33.1} \\
mhr\_Cyrl & 36.0 & \textbf{38.5} & min\_Latn & \textbf{49.9} & 40.7 & miq\_Latn & \textbf{52.2} & 52.2 & mkd\_Cyrl & \textbf{71.2} & 70.3 \\
mlt\_Latn & \textbf{50.7} & 50.7 & mos\_Latn & 40.3 & \textbf{41.2} & mps\_Latn & \textbf{57.1} & 53.1 & mri\_Latn & 50.9 & \textbf{52.6} \\
    \bottomrule
    \end{tabular}
}
    \caption{F1 scores of \frameworkname on \textbf{Taxi1500} using English and the closest donor language as source (Part I).}\label{tab:taxi_donor_table1}
\end{table*}

\begin{table*}
\centering
\setlength{\tabcolsep}{0.7mm}{}
\resizebox{\textwidth}{!}{
    \begin{tabular}{lrr|lrr|lrr|lrr}
    \toprule
    Language & English & Closest donor &     Language & English & Closest donor &     Language & English & Closest donor &     Language & English & Closest donor \\
    \midrule
msa\_Latn & 41.7 & \textbf{42.0} & mwm\_Latn & \textbf{55.1} & 55.0 & mxv\_Latn & \textbf{29.6} & 27.4 & mya\_Mymr & \textbf{54.4} & 53.4 \\
mzh\_Latn & 39.7 & \textbf{45.1} & nan\_Latn & 31.5 & \textbf{31.8} & naq\_Latn & 41.7 & \textbf{43.7} & nav\_Latn & 21.1 & \textbf{29.5} \\
nch\_Latn & \textbf{44.0} & 36.6 & ncj\_Latn & 38.6 & \textbf{39.1} & ndc\_Latn & 34.7 & \textbf{36.6} & nde\_Latn & 45.7 & \textbf{49.8} \\
nds\_Latn & \textbf{49.6} & 44.0 & nep\_Deva & 68.0 & \textbf{72.1} & ngu\_Latn & 43.4 & \textbf{48.2} & nld\_Latn & \textbf{61.1} & 53.7 \\
nnb\_Latn & 40.7 & \textbf{46.1} & nno\_Latn & \textbf{63.1} & 63.1 & nob\_Latn & 57.2 & \textbf{58.2} & nor\_Latn & 56.4 & \textbf{57.8} \\
nse\_Latn & 45.9 & \textbf{48.5} & nso\_Latn & \textbf{48.6} & 48.6 & nya\_Latn & \textbf{56.0} & 47.4 & nyn\_Latn & 43.0 & \textbf{44.1} \\
nzi\_Latn & 33.0 & \textbf{33.8} & ori\_Orya & \textbf{67.3} & 67.3 & ory\_Orya & 66.9 & \textbf{70.7} & oss\_Cyrl & 55.5 & \textbf{57.5} \\
pag\_Latn & \textbf{55.5} & 52.5 & pam\_Latn & \textbf{42.0} & 37.8 & pan\_Guru & \textbf{64.1} & 64.1 & pap\_Latn & \textbf{65.6} & 59.8 \\
pcm\_Latn & \textbf{66.1} & 65.9 & pdt\_Latn & \textbf{60.0} & 56.5 & pes\_Arab & \textbf{69.0} & 69.0 & pis\_Latn & 64.3 & \textbf{65.0} \\
plt\_Latn & 46.8 & \textbf{52.9} & poh\_Latn & 44.3 & \textbf{45.5} & pol\_Latn & 64.8 & \textbf{65.1} & pon\_Latn & 50.5 & \textbf{52.2} \\
prk\_Latn & 52.9 & \textbf{53.0} & prs\_Arab & 69.2 & \textbf{70.0} & pxm\_Latn & 34.5 & \textbf{41.5} & qub\_Latn & 51.5 & \textbf{56.3} \\
qug\_Latn & \textbf{65.0} & 61.3 & quh\_Latn & \textbf{66.7} & 58.8 & quw\_Latn & 55.9 & \textbf{56.0} & quy\_Latn & 65.5 & \textbf{67.7} \\
qvi\_Latn & \textbf{62.0} & 58.5 & rap\_Latn & 48.9 & \textbf{49.3} & rar\_Latn & 48.9 & \textbf{51.9} & rmy\_Latn & 45.4 & \textbf{49.1} \\
rop\_Latn & \textbf{56.6} & 54.7 & rug\_Latn & 53.8 & \textbf{55.1} & run\_Latn & 48.0 & \textbf{55.2} & rus\_Cyrl & \textbf{68.1} & 68.1 \\
sah\_Cyrl & 55.1 & \textbf{57.6} & sba\_Latn & 39.1 & \textbf{41.4} & seh\_Latn & 45.0 & \textbf{46.7} & sin\_Sinh & 64.1 & \textbf{66.9} \\
slv\_Latn & \textbf{63.8} & 60.7 & sme\_Latn & \textbf{42.8} & 37.6 & smo\_Latn & \textbf{60.8} & 54.2 & sna\_Latn & 42.6 & \textbf{44.9} \\
som\_Latn & 33.9 & \textbf{35.5} & sop\_Latn & \textbf{36.4} & 36.0 & sot\_Latn & 43.5 & \textbf{45.5} & spa\_Latn & \textbf{64.2} & 64.2 \\
srm\_Latn & 48.1 & \textbf{48.4} & srn\_Latn & \textbf{63.7} & 62.8 & srp\_Latn & 64.9 & \textbf{65.2} & ssw\_Latn & \textbf{43.7} & 37.7 \\
suz\_Deva & \textbf{58.0} & 57.8 & swe\_Latn & \textbf{66.8} & 65.3 & swh\_Latn & \textbf{59.8} & 59.8 & sxn\_Latn & \textbf{46.6} & 40.2 \\
tat\_Cyrl & 62.2 & \textbf{68.2} & tbz\_Latn & 36.4 & \textbf{39.5} & tca\_Latn & 43.3 & \textbf{50.3} & tdt\_Latn & \textbf{60.3} & 55.1 \\
teo\_Latn & \textbf{23.7} & 23.1 & tgk\_Cyrl & \textbf{60.9} & 60.9 & tgl\_Latn & 56.7 & \textbf{58.7} & tha\_Thai & \textbf{63.8} & 63.8 \\
tir\_Ethi & \textbf{50.1} & 50.1 & tlh\_Latn & \textbf{65.0} & 65.0 & tob\_Latn & 43.3 & \textbf{50.4} & toh\_Latn & 37.1 & \textbf{39.0} \\
toj\_Latn & \textbf{36.6} & 34.1 & ton\_Latn & 47.3 & \textbf{51.5} & top\_Latn & \textbf{21.9} & 21.3 & tpi\_Latn & 63.8 & \textbf{67.6} \\
tsn\_Latn & 39.8 & \textbf{44.1} & tsz\_Latn & 40.4 & \textbf{41.0} & tuc\_Latn & \textbf{57.4} & 56.9 & tui\_Latn & \textbf{43.7} & 43.7 \\
tum\_Latn & \textbf{47.6} & 43.2 & tur\_Latn & \textbf{62.1} & 62.1 & twi\_Latn & \textbf{41.4} & 38.9 & tyv\_Cyrl & 59.8 & \textbf{60.3} \\
tzo\_Latn & \textbf{39.5} & 39.5 & udm\_Cyrl & 49.6 & \textbf{49.9} & ukr\_Cyrl & \textbf{62.4} & 62.2 & uzb\_Latn & 53.5 & \textbf{57.7} \\
ven\_Latn & 41.9 & \textbf{48.6} & vie\_Latn & 62.4 & \textbf{65.4} & wal\_Latn & \textbf{48.9} & 42.7 & war\_Latn & 47.7 & \textbf{54.5} \\
wol\_Latn & \textbf{37.2} & 33.9 & xav\_Latn & \textbf{25.5} & 23.7 & xho\_Latn & \textbf{44.9} & 44.4 & yan\_Latn & 50.3 & \textbf{53.5} \\
yap\_Latn & 42.8 & \textbf{42.9} & yom\_Latn & \textbf{37.6} & 34.1 & yor\_Latn & \textbf{41.8} & 35.4 & yua\_Latn & 40.1 & \textbf{43.2} \\
zai\_Latn & \textbf{42.6} & 41.4 & zho\_Hani & \textbf{60.7} & 60.7 & zlm\_Latn & \textbf{68.4} & 65.5 & zom\_Latn & \textbf{44.6} & 44.4 \\
zul\_Latn & 51.9 & \textbf{52.2} \\

    \bottomrule
    \end{tabular}
}
    \caption{F1 scores of \frameworkname on \textbf{Taxi1500} using English and the closest donor language as source (Part II).}\label{tab:taxi_donor_table2}
\end{table*}

%% file: sib200_donor.tex
\begin{table*}
\centering
\setlength{\tabcolsep}{0.7mm}{}
\resizebox{\textwidth}{!}{
    \begin{tabular}{lrr|lrr|lrr|lrr}
    \toprule
    Language & English & Closest donor &     Language & English & Closest donor &     Language & English & Closest donor &     Language & English & Closest donor \\
    \midrule
ace\_Latn & 69.9 & \textbf{72.4} & acm\_Arab & 80.6 & \textbf{81.4} & afr\_Latn & 81.4 & \textbf{81.8} & ajp\_Arab & 81.4 & \textbf{83.0} \\
als\_Latn & \textbf{82.3} & 82.3 & amh\_Ethi & \textbf{72.6} & 72.6 & apc\_Arab & 81.7 & \textbf{83.2} & arb\_Arab & \textbf{81.5} & 81.5 \\
arz\_Arab & 82.1 & \textbf{84.4} & asm\_Beng & \textbf{83.0} & 83.0 & ast\_Latn & 87.1 & \textbf{87.6} & ayr\_Latn & 48.6 & \textbf{51.1} \\
azj\_Latn & \textbf{86.5} & 84.0 & bak\_Cyrl & 84.3 & \textbf{86.5} & bam\_Latn & \textbf{46.5} & 42.2 & ban\_Latn & 79.5 & \textbf{81.3} \\
bem\_Latn & \textbf{61.1} & 51.4 & ben\_Beng & 83.7 & \textbf{84.0} & bjn\_Latn & 75.9 & \textbf{77.9} & bod\_Tibt & 65.7 & \textbf{71.0} \\
bul\_Cyrl & 86.3 & \textbf{86.6} & cat\_Latn & \textbf{85.7} & 85.2 & ceb\_Latn & 81.2 & \textbf{83.2} & ces\_Latn & \textbf{86.3} & 85.6 \\
ckb\_Arab & \textbf{80.0} & 76.8 & crh\_Latn & \textbf{76.8} & 75.7 & cym\_Latn & 73.6 & \textbf{76.6} & dan\_Latn & 85.0 & \textbf{86.0} \\
dyu\_Latn & \textbf{43.6} & 42.4 & dzo\_Tibt & \textbf{68.2} & 59.8 & ell\_Grek & \textbf{79.5} & 78.8 & eng\_Latn & \textbf{88.9} & 88.9 \\
est\_Latn & \textbf{78.9} & 78.1 & eus\_Latn & 78.8 & \textbf{80.7} & ewe\_Latn & \textbf{49.9} & 46.7 & fao\_Latn & \textbf{84.4} & 83.6 \\
fin\_Latn & 80.9 & \textbf{81.5} & fon\_Latn & \textbf{40.8} & 38.1 & fra\_Latn & \textbf{87.8} & 87.8 & fur\_Latn & 77.4 & \textbf{77.9} \\
gle\_Latn & 61.5 & \textbf{64.4} & glg\_Latn & \textbf{87.6} & 87.6 & grn\_Latn & 71.6 & \textbf{73.2} & guj\_Gujr & 82.1 & \textbf{83.4} \\
hau\_Latn & 59.3 & \textbf{64.2} & heb\_Hebr & 76.8 & \textbf{80.2} & hin\_Deva & \textbf{82.8} & 82.8 & hne\_Deva & 77.9 & \textbf{79.5} \\
hun\_Latn & 86.6 & \textbf{87.5} & hye\_Armn & \textbf{81.3} & 80.3 & ibo\_Latn & \textbf{71.4} & 71.3 & ilo\_Latn & 76.1 & \textbf{76.7} \\
isl\_Latn & 78.0 & \textbf{78.3} & ita\_Latn & 86.4 & \textbf{87.5} & jav\_Latn & \textbf{79.9} & 79.7 & jpn\_Jpan & \textbf{86.8} & 86.8 \\
kac\_Latn & \textbf{48.9} & 46.6 & kam\_Latn & 45.8 & \textbf{48.3} & kan\_Knda & 82.9 & \textbf{83.0} & kat\_Geor & \textbf{83.7} & 81.0 \\
kbp\_Latn & \textbf{42.8} & 42.2 & kea\_Latn & \textbf{73.1} & 73.1 & khm\_Khmr & \textbf{82.7} & 82.7 & kik\_Latn & 55.1 & \textbf{56.7} \\
kir\_Cyrl & 79.3 & \textbf{80.1} & kmb\_Latn & \textbf{46.2} & 42.6 & kmr\_Latn & \textbf{69.8} & 68.9 & kon\_Latn & \textbf{65.2} & 63.4 \\
lao\_Laoo & \textbf{83.4} & 82.9 & lij\_Latn & \textbf{76.4} & 74.9 & lim\_Latn & \textbf{74.1} & 73.0 & lin\_Latn & 68.2 & \textbf{73.3} \\
lmo\_Latn & 77.0 & \textbf{78.3} & ltz\_Latn & \textbf{76.4} & 76.4 & lua\_Latn & \textbf{54.4} & 54.3 & lug\_Latn & \textbf{58.2} & 55.8 \\
lus\_Latn & \textbf{64.8} & 64.8 & lvs\_Latn & \textbf{83.2} & 83.0 & mai\_Deva & \textbf{82.9} & 82.1 & mal\_Mlym & \textbf{79.8} & 79.3 \\
min\_Latn & 76.7 & \textbf{79.8} & mkd\_Cyrl & \textbf{83.6} & 82.8 & mlt\_Latn & \textbf{81.3} & 81.3 & mos\_Latn & \textbf{44.7} & 40.9 \\
mya\_Mymr & \textbf{80.5} & 78.8 & nld\_Latn & 85.1 & \textbf{86.4} & nno\_Latn & \textbf{86.0} & 86.0 & nob\_Latn & \textbf{84.8} & 84.4 \\
nso\_Latn & \textbf{57.6} & 57.6 & nya\_Latn & 69.2 & \textbf{70.9} & oci\_Latn & \textbf{85.0} & 84.1 & ory\_Orya & 78.6 & \textbf{79.0} \\
pan\_Guru & \textbf{76.4} & 76.4 & pap\_Latn & 76.9 & \textbf{78.1} & pes\_Arab & \textbf{87.5} & 87.3 & plt\_Latn & 67.5 & \textbf{69.3} \\
por\_Latn & 85.3 & \textbf{86.8} & prs\_Arab & 85.0 & \textbf{85.5} & quy\_Latn & \textbf{62.6} & 59.7 & ron\_Latn & 84.0 & \textbf{84.4} \\
rus\_Cyrl & \textbf{86.8} & 86.8 & sag\_Latn & \textbf{51.3} & 50.2 & san\_Deva & 72.9 & \textbf{76.6} & sat\_Olck & \textbf{56.4} & 53.5 \\
sin\_Sinh & \textbf{82.7} & 82.7 & slk\_Latn & \textbf{85.4} & 85.1 & slv\_Latn & 84.2 & \textbf{87.4} & smo\_Latn & 74.2 & \textbf{75.3} \\
snd\_Arab & \textbf{70.4} & 70.4 & som\_Latn & 58.9 & \textbf{61.1} & sot\_Latn & \textbf{64.1} & 63.2 & spa\_Latn & \textbf{84.4} & 84.4 \\
srp\_Cyrl & 84.8 & \textbf{85.0} & ssw\_Latn & 64.1 & \textbf{65.2} & sun\_Latn & 82.6 & \textbf{85.2} & swe\_Latn & 84.2 & \textbf{86.2} \\
szl\_Latn & \textbf{72.4} & 72.4 & tam\_Taml & \textbf{81.2} & 81.2 & tat\_Cyrl & \textbf{83.6} & 83.6 & tel\_Telu & 84.0 & \textbf{85.4} \\
tgl\_Latn & \textbf{82.1} & 81.7 & tha\_Thai & 85.4 & \textbf{85.7} & tir\_Ethi & \textbf{60.3} & 60.3 & tpi\_Latn & \textbf{80.3} & 75.7 \\
tso\_Latn & 57.3 & \textbf{60.3} & tuk\_Latn & 78.1 & \textbf{78.5} & tum\_Latn & 65.4 & \textbf{68.5} & tur\_Latn & \textbf{80.4} & 80.4 \\
uig\_Arab & \textbf{75.5} & 75.5 & ukr\_Cyrl & \textbf{84.3} & 83.8 & umb\_Latn & 41.0 & \textbf{46.5} & urd\_Arab & 79.1 & \textbf{80.6} \\
vie\_Latn & \textbf{86.2} & 83.9 & war\_Latn & 80.7 & \textbf{81.3} & wol\_Latn & \textbf{50.5} & 46.4 & xho\_Latn & \textbf{60.1} & 59.8 \\
zho\_Hans & \textbf{89.6} & 89.2 & zho\_Hant & \textbf{88.8} & 88.8 & zsm\_Latn & \textbf{86.4} & 86.0 & zul\_Latn & 68.1 & \textbf{69.8} \\

    \bottomrule
    \end{tabular}
}
    \caption{F1 scores of \frameworkname on \textbf{SIB200}. using English and the closest donor language as source.}\label{tab:sib200_donor_table1}
\end{table*}

%% file: ner_donor.tex
\begin{table*}
\centering
\setlength{\tabcolsep}{0.7mm}{}
\resizebox{\textwidth}{!}{
    \begin{tabular}{lrr|lrr|lrr|lrr}
    \toprule
    Language & English & Closest donor &     Language & English & Closest donor &     Language & English & Closest donor &     Language & English & Closest donor \\
    \midrule
ace\_Latn & 41.5 & \textbf{56.9} & afr\_Latn & 75.8 & \textbf{80.3} & als\_Latn & \textbf{80.9} & 80.9 & amh\_Ethi & \textbf{39.7} & 39.7 \\
arg\_Latn & 82.2 & \textbf{88.8} & arz\_Arab & 55.1 & \textbf{82.6} & asm\_Beng & \textbf{69.0} & 45.9 & ast\_Latn & 84.6 & \textbf{85.8} \\
aze\_Latn & 65.0 & \textbf{74.0} & bak\_Cyrl & 62.5 & \textbf{72.2} & bar\_Latn & \textbf{68.2} & 62.8 & bel\_Cyrl & 74.9 & \textbf{79.7} \\
bih\_Deva & 56.2 & \textbf{67.6} & bod\_Tibt & 35.2 & \textbf{35.7} & bos\_Latn & 70.1 & \textbf{75.2} & bre\_Latn & 63.3 & \textbf{66.0} \\
cat\_Latn & 83.8 & \textbf{85.1} & cbk\_Latn & \textbf{53.7} & 48.9 & ceb\_Latn & \textbf{56.0} & 26.8 & ces\_Latn & \textbf{77.9} & 69.6 \\
chv\_Cyrl & 73.6 & \textbf{84.3} & ckb\_Arab & \textbf{76.0} & 60.6 & cos\_Latn & \textbf{63.0} & 61.9 & crh\_Latn & 52.7 & \textbf{59.4} \\
cym\_Latn & 61.7 & \textbf{62.1} & dan\_Latn & \textbf{81.4} & 81.3 & deu\_Latn & \textbf{74.6} & 74.6 & diq\_Latn & 54.0 & \textbf{72.2} \\
ell\_Grek & 71.9 & \textbf{72.0} & eml\_Latn & \textbf{41.3} & 41.3 & eng\_Latn & \textbf{83.5} & 83.5 & epo\_Latn & \textbf{68.3} & 68.3 \\
eus\_Latn & 60.9 & \textbf{65.1} & ext\_Latn & 44.2 & \textbf{48.6} & fao\_Latn & 68.7 & \textbf{79.2} & fas\_Arab & \textbf{55.0} & 53.6 \\
fra\_Latn & \textbf{76.5} & 76.5 & frr\_Latn & \textbf{52.0} & 52.0 & fry\_Latn & \textbf{74.6} & 73.9 & fur\_Latn & \textbf{58.2} & 54.0 \\
gle\_Latn & \textbf{72.6} & 69.6 & glg\_Latn & 80.7 & \textbf{86.1} & grn\_Latn & 55.1 & \textbf{59.8} & guj\_Gujr & \textbf{61.2} & 61.0 \\
heb\_Hebr & 52.0 & \textbf{52.9} & hin\_Deva & \textbf{69.4} & 69.4 & hrv\_Latn & 77.2 & \textbf{79.8} & hsb\_Latn & \textbf{74.3} & 69.7 \\
hye\_Armn & 53.0 & \textbf{62.2} & ibo\_Latn & 58.1 & \textbf{58.4} & ido\_Latn & \textbf{82.6} & 81.5 & ilo\_Latn & \textbf{80.0} & 74.9 \\
ind\_Latn & \textbf{67.6} & 67.6 & isl\_Latn & 70.1 & \textbf{75.4} & ita\_Latn & 78.2 & \textbf{79.5} & jav\_Latn & 56.0 & \textbf{86.4} \\
jpn\_Jpan & \textbf{22.0} & 22.0 & kan\_Knda & 57.5 & \textbf{61.8} & kat\_Geor & \textbf{68.7} & 60.1 & kaz\_Cyrl & 50.5 & \textbf{57.1} \\
kin\_Latn & \textbf{69.6} & 67.3 & kir\_Cyrl & 44.3 & \textbf{60.9} & kor\_Hang & 50.4 & \textbf{51.2} & ksh\_Latn & \textbf{59.7} & 51.4 \\
lat\_Latn & 71.9 & \textbf{81.4} & lav\_Latn & \textbf{74.4} & 69.0 & lij\_Latn & 45.2 & \textbf{54.2} & lim\_Latn & \textbf{69.3} & 61.2 \\
lit\_Latn & 74.2 & \textbf{76.1} & lmo\_Latn & \textbf{73.6} & 65.5 & ltz\_Latn & \textbf{67.9} & 67.9 & lzh\_Hani & \textbf{14.8} & 14.8 \\
mar\_Deva & 62.5 & \textbf{76.6} & mhr\_Cyrl & 60.6 & \textbf{72.3} & min\_Latn & 42.6 & \textbf{57.5} & mkd\_Cyrl & 72.2 & \textbf{73.1} \\
mlt\_Latn & \textbf{75.9} & 75.9 & mon\_Cyrl & \textbf{68.7} & 60.9 & mri\_Latn & \textbf{50.0} & 47.0 & msa\_Latn & 67.6 & \textbf{73.0} \\
mya\_Mymr & 55.3 & \textbf{56.3} & mzn\_Arab & 43.3 & \textbf{47.2} & nan\_Latn & \textbf{88.1} & 36.6 & nap\_Latn & \textbf{63.0} & 55.3 \\
nep\_Deva & 56.9 & \textbf{60.4} & nld\_Latn & \textbf{80.8} & 80.0 & nno\_Latn & \textbf{77.6} & 77.6 & nor\_Latn & 77.9 & \textbf{80.4} \\
ori\_Orya & \textbf{34.2} & 34.2 & oss\_Cyrl & 50.6 & \textbf{59.1} & pan\_Guru & \textbf{51.5} & 51.5 & pms\_Latn & \textbf{80.9} & 78.4 \\
pol\_Latn & \textbf{77.7} & 71.1 & por\_Latn & 78.9 & \textbf{84.9} & pus\_Arab & 42.6 & \textbf{45.3} & que\_Latn & \textbf{70.4} & 55.5 \\
ron\_Latn & \textbf{77.8} & 75.5 & rus\_Cyrl & \textbf{67.5} & 67.5 & sah\_Cyrl & 71.9 & \textbf{77.9} & san\_Deva & 38.4 & \textbf{53.4} \\
sco\_Latn & \textbf{86.4} & 84.5 & sgs\_Latn & 66.4 & \textbf{69.8} & sin\_Sinh & \textbf{53.0} & 51.2 & slk\_Latn & \textbf{76.4} & 55.9 \\
snd\_Arab & \textbf{41.8} & 41.8 & som\_Latn & \textbf{57.5} & 56.2 & spa\_Latn & \textbf{77.6} & 77.6 & sqi\_Latn & 76.8 & \textbf{78.7} \\
sun\_Latn & 50.8 & \textbf{75.1} & swa\_Latn & \textbf{71.8} & 71.8 & swe\_Latn & \textbf{70.9} & 65.8 & szl\_Latn & \textbf{70.9} & 70.9 \\
tat\_Cyrl & 63.8 & \textbf{76.5} & tel\_Telu & 48.1 & \textbf{49.0} & tgk\_Cyrl & \textbf{68.4} & 68.4 & tgl\_Latn & 71.9 & \textbf{73.7} \\
tuk\_Latn & 54.4 & \textbf{57.3} & tur\_Latn & \textbf{77.1} & 77.1 & uig\_Arab & 47.7 & \textbf{62.3} & ukr\_Cyrl & 76.6 & \textbf{85.3} \\
uzb\_Latn & 73.2 & \textbf{76.0} & vec\_Latn & 68.0 & \textbf{75.1} & vep\_Latn & \textbf{72.0} & 63.0 & vie\_Latn & \textbf{72.3} & 49.7 \\
vol\_Latn & \textbf{61.0} & 36.5 & war\_Latn & \textbf{64.9} & 56.1 & wuu\_Hani & 35.7 & \textbf{66.7} & xmf\_Geor & \textbf{69.3} & 55.7 \\
yor\_Latn & \textbf{69.3} & 41.7 & yue\_Hani & 25.7 & \textbf{73.5} & zea\_Latn & 62.9 & \textbf{75.4} & zho\_Hani & \textbf{25.2} & 25.2 \\

    \bottomrule
    \end{tabular}
}
    \caption{F1 scores of \frameworkname on \textbf{NER} using English and the closest donor language as source.}\label{tab:ner_donor_table1}
\end{table*}

%% file: pos_donor.tex
\begin{table*}
\centering
\setlength{\tabcolsep}{0.7mm}{}
\resizebox{\textwidth}{!}{
    \begin{tabular}{lrr|lrr|lrr|lrr}
    \toprule
    Language & English & Closest donor &     Language & English & Closest donor &     Language & English & Closest donor &     Language & English & Closest donor \\
    \midrule
afr\_Latn & \textbf{88.5} & 79.5 & ajp\_Arab & \textbf{71.1} & 41.9 & aln\_Latn & \textbf{53.4} & 45.1 & amh\_Ethi & \textbf{66.8} & 66.8 \\
bam\_Latn & \textbf{43.0} & 31.2 & bel\_Cyrl & 86.4 & \textbf{93.8} & ben\_Beng & \textbf{87.5} & 80.2 & bre\_Latn & 61.1 & \textbf{62.3} \\
cat\_Latn & 86.8 & \textbf{95.8} & ceb\_Latn & \textbf{66.7} & 32.5 & ces\_Latn & \textbf{85.4} & 73.3 & cym\_Latn & \textbf{65.5} & 60.4 \\
deu\_Latn & \textbf{88.2} & 88.2 & ell\_Grek & \textbf{84.9} & 75.5 & eng\_Latn & \textbf{96.0} & 96.0 & est\_Latn & \textbf{84.7} & 77.4 \\
fao\_Latn & \textbf{88.7} & 67.5 & fas\_Arab & \textbf{72.2} & 69.1 & fin\_Latn & \textbf{82.2} & 75.8 & fra\_Latn & \textbf{85.8} & 85.8 \\
gle\_Latn & 64.6 & \textbf{65.5} & glg\_Latn & 83.6 & \textbf{87.8} & glv\_Latn & 51.9 & \textbf{57.8} & grc\_Grek & \textbf{71.6} & 71.6 \\
gsw\_Latn & \textbf{82.7} & 82.7 & hbo\_Hebr & \textbf{38.9} & 37.4 & heb\_Hebr & 67.9 & \textbf{69.3} & hin\_Deva & \textbf{77.2} & 77.2 \\
hsb\_Latn & \textbf{83.7} & 73.4 & hun\_Latn & \textbf{82.2} & 42.0 & hye\_Armn & \textbf{85.1} & 84.9 & hyw\_Armn & \textbf{83.0} & 56.8 \\
isl\_Latn & \textbf{82.7} & 81.2 & ita\_Latn & 88.9 & \textbf{92.4} & jav\_Latn & 75.4 & \textbf{78.8} & jpn\_Jpan & \textbf{33.1} & 33.1 \\
kmr\_Latn & \textbf{76.6} & 61.6 & kor\_Hang & \textbf{52.7} & 45.3 & lat\_Latn & 72.8 & \textbf{74.2} & lav\_Latn & \textbf{83.7} & 78.4 \\
lit\_Latn & \textbf{82.1} & 80.7 & lzh\_Hani & \textbf{24.5} & 24.5 & mal\_Mlym & \textbf{86.0} & 52.1 & mar\_Deva & \textbf{84.1} & 81.7 \\
myv\_Cyrl & \textbf{65.9} & 58.4 & nap\_Latn & \textbf{82.4} & 70.6 & nds\_Latn & \textbf{79.1} & 34.0 & nld\_Latn & \textbf{88.2} & 82.2 \\
pcm\_Latn & \textbf{58.2} & 48.1 & pol\_Latn & 84.2 & \textbf{89.1} & por\_Latn & 87.9 & \textbf{92.0} & quc\_Latn & \textbf{63.3} & 52.6 \\
rus\_Cyrl & \textbf{88.7} & 88.7 & sah\_Cyrl & 74.2 & \textbf{74.5} & san\_Deva & 25.5 & \textbf{32.7} & sin\_Sinh & \textbf{56.2} & 34.4 \\
slv\_Latn & 77.6 & \textbf{79.0} & sme\_Latn & \textbf{74.8} & 60.6 & spa\_Latn & \textbf{87.8} & 87.8 & sqi\_Latn & \textbf{77.5} & 72.7 \\
swe\_Latn & \textbf{92.7} & 83.2 & tam\_Taml & \textbf{74.6} & 74.6 & tat\_Cyrl & \textbf{72.4} & 70.9 & tel\_Telu & \textbf{80.9} & 55.9 \\
tha\_Thai & \textbf{58.3} & 27.5 & tur\_Latn & \textbf{71.2} & 71.2 & uig\_Arab & \textbf{68.2} & 48.3 & ukr\_Cyrl & 85.6 & \textbf{91.7} \\
vie\_Latn & \textbf{68.4} & 32.4 & wol\_Latn & \textbf{61.6} & 57.4 & xav\_Latn & \textbf{16.7} & 11.2 & yor\_Latn & \textbf{62.7} & 46.5 \\
zho\_Hani & \textbf{47.4} & 47.4 \\

    \bottomrule
    \end{tabular}
}
    \caption{F1 scores of \frameworkname on \textbf{POS} using English and the closest donor language as source.}\label{tab:pos_donor_table1}
\end{table*}